\documentclass[]{fairmeta}

\usepackage{enumitem}
\usepackage{float}
\usepackage{afterpage}
\usepackage{wrapfig}
\usepackage{tabularx}
\usepackage{longtable}
\usepackage{multirow}
\usepackage{booktabs}
\usepackage{tikz}
\usetikzlibrary{arrows.meta, positioning, shapes.geometric, calc}
\usepackage{amssymb}
\usepackage{amsmath}
\usepackage{graphicx}
\usepackage{subcaption}
\usepackage{hyperref}
\usepackage{url}
\usepackage{microtype}
\usepackage{xcolor}
\newtcolorbox{promptquote}[1]{
  breakable,
  colback=black!2, colframe=black!30,
  colbacktitle=black!8, coltitle=black,
  fonttitle=\sffamily\bfseries\small,
  fontupper=\sffamily\small,
  title={#1}, title after break={#1 (continued)},
  boxrule=0.4pt, arc=1mm,
  left=7pt, right=7pt, top=6pt, bottom=6pt,
  before skip=10pt, after skip=10pt,
  before upper={\raggedright\setlength{\parindent}{0pt}\setlength{\parskip}{2pt}}
}

\definecolor{metabg}{HTML}{F1F4F7}
\DeclareFontShape{T1}{optimistic}{m}{it}{<-> s * [0.88] assets-final/optimistic}{}  %
\DisableLigatures[f]{family=optimistic}  %
\newcommand{\optimisticfont}{\fontfamily{optimistic}\selectfont\renewcommand{\sfdefault}{optimistic}}

\newcommand{\warn}[1]{}

\newcommand{\hone}{\texttt{h1}}
\newcommand{\ahmad}[1]{\textcolor{red}{}}

\title{
\hone{}
: A Taste of the Bitter Lesson in Automated Theorem Proving
}

\author[1,2,\dagger]{Skander Moalla}
\author[1,3]{Ahmad Rammal}
\author[1,4,\dagger]{Niket Patel}
\author[1]{R\'emi Munos}
\author[1,4,\dagger]{Julia Kempe}

\affiliation[1]{FAIR at Meta}
\affiliation[2]{EPFL}
\affiliation[3]{CERMICS, ENPC, Institut Polytechnique de Paris}
\affiliation[4]{New York University}
\contribution[\dagger]{Work done at Meta}

\abstract{
While progress in informal mathematics has come largely from scaling the chain-of-thought and agentic reasoning of frontier language models, formal theorem proving has increasingly relied on human-designed workflows around these models, such as planning, decomposition, and orchestrating specialized provers.
This raises a bitter-lesson question: can a simple agentic loop that scales test-time compute match these workflows?
The answer matters as formal verification must scale to keep up with frontier models producing research mathematics, yet new workflows rarely settle it, as their gains are often confounded with stronger models or more compute.
We study this question with \hone{}, a simple harness that gives a model an efficient interface to Lean~4 and keeps it working through verifier feedback and self-reflection, without imposing a decomposition or search structure.
\hone{} reaches state-of-the-art results on PutnamBench, \mbox{FATE-X}, and FormalConjectures100-Solved, settles eight open problems in FormalConjectures, and, at matched models, matches or outperforms specialized workflows such as Goedel-Architect and AlphaProof Nexus, surprisingly showing that a single attempt (pass@1) can often scale to solve open problems.
Through ablations and matched-model comparisons with Claude Code and AxProverBase, we find that simplicity alone is not enough: an efficient Lean interface largely determines cost and solve rate.
We thus recommend that future work evaluate frontier models with efficient execution feedback before prescribing workflows, at the risk of bitter lessons.
}

\date{October 30, 2026}
\correspondence{Skander Moalla at \email{skander.moalla@epfl.ch}}

\begin{document}
\maketitle
\setcounter{topnumber}{1}
\setcounter{totalnumber}{1}

\vspace{-.8\baselineskip}
\begin{figure}[H]
\captionsetup{skip=7pt}
\centering
\makebox[\textwidth][c]{\resizebox{\textwidth}{!}{%
\includegraphics[width=155bp]{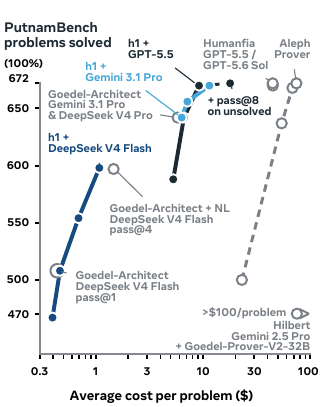}\hspace{4.881515bp}%
\includegraphics[width=155bp]{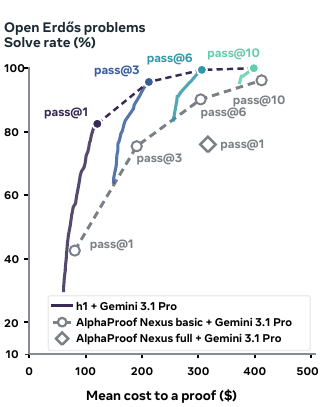}\hspace{-1.815831bp}%
\includegraphics[width=155bp]{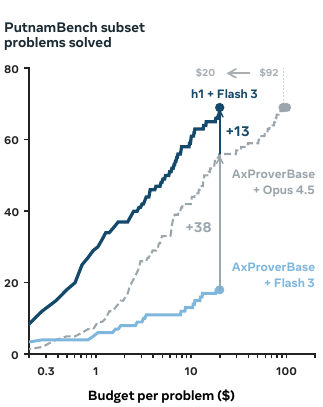}}}
\caption{\textbf{(Left) Simple scaling Pareto-dominates specialized systems.} On PutnamBench (672 competition problems), \hone{} with DeepSeek~V4~Flash Pareto-dominates Goedel-Architect's best of four runs (pass@4) seeded with natural-language proofs.
With GPT-5.5, it solves every solvable problem, as Humanfia and Aleph Prover do, at lower cost.
\textbf{(Middle) Longer attempts can outperform repetition.} On five open Erd\H{o}s problems, a single \hone{} attempt (pass@1) beats three attempts of AlphaProof Nexus's basic agent (APN basic, pass@3) at lower mean cost per success, can be cheaper than repeating \hone{} itself, and Pareto-dominates the full APN system.
\textbf{(Right) Simple harness design is important for scaling efficiency and effectiveness.} On a 100-problem PutnamBench subset, Gemini~3~Flash trails Opus~4.5 in the AxProverBase loop but overtakes it in \hone{} at matched budget through more efficient tool use and caching.
}
\label{fig:teaser}
\end{figure}

\addtocontents{toc}{\protect\setcounter{tocdepth}{-1}}
\section{Introduction}
\label{sec:intro}

The bitter lesson predicts that general methods which scale with computation ultimately outperform approaches built around human domain knowledge~\citep{sutton2019bitterlesson}.
However, while progress in informal mathematics has come from scaling frontier models' chain-of-thought and agentic reasoning~\citep{openai2024o1,deepseekr1,rstar2agent,retool}, progress in formal proving has increasingly relied on human-designed workflows around those models.
Indeed, formal proof assistants such as Lean~\citep{moura2021lean4} express mathematical statements and proofs in a precise language that can be checked mechanically.
This includes intermediate proof steps, which naturally supports structured search that can be encoded as part of a workflow in a \emph{harness}: the tools, context, and control loop around a model.
Recent formal proving harnesses exploit this structure by orchestrating specialized provers or prescribing planning, decomposition, and proof refinement~\citep{hilbert2025,goedelarchitect2026,merlean2026prover,alephprover2026,opengauss2026}.
Only very recently has interest grown in \emph{simple} loops that use the harness \emph{only} to provide Lean feedback and sustain work~\citep{axprover,alphaproofnexus2026,humanfia2026,leanstral2026}.

Formal mathematics thus raises a bitter-lesson question for harness design: with frontier models already capable of reasoning and agentic test-time scaling, can a simple test-time scaling loop match specialized workflows?
And what should a simple harness provide to make that scaling effective and efficient?
Do these simple methods offer better performance for the same cost?
These questions matter increasingly as frontier models generate research mathematics whose informal arguments are difficult to verify.
Indeed, scaling automated theorem proving (ATP) to produce mechanically checked proofs has become a necessary component of automated discovery~\citep{alphaproofnexus2026,openai2026tenadvances,openai2026navierstokes,patel2026learningdiscoverinterestingmathematics}.
Yet new workflows are often presented alongside stronger models and more compute, or compared to baselines that scale poorly.
This leaves unclear whether their gains hold against an efficient simple loop given enough test-time compute.

We tackle these questions with {\hone{}}, a simple harness with an efficient interface to Lean 4 that imposes no decomposition or search structure.
This loop keeps a model working until success or a cost cap stops it by iterating between a multi-turn block of reasoning and tool-calling, submission to a formal validator, and a reflection that compacts the transcript into a new context for the next iteration.
Each additional iteration is akin to appending ``Wait'' to a model's thinking in s1~\citep{muennighoff2025s1}, extending the computation available at test time.
We evaluate cost--success curves across benchmarks and open problems, accounting for inference cost and matching models where possible.
Our findings are threefold:

\begin{itemize}[leftmargin=5mm,nosep,topsep=4pt,itemsep=4pt]
  \item \textbf{A bitter lesson for specialized harnesses: simple test-time scaling reaches frontier results without prescribed workflows} (Sections~\ref{sec:benchmarks} and~\ref{sec:complex}).
  A simple loop often matches or Pareto-dominates specialized systems at matched models and comparable budgets, while reaching state-of-the-art results on PutnamBench~\citep{tsoukalas2024putnambench}, FATE-X~\citep{fate2025}, and FormalConjectures100-Solved~\citep{formalconjectures} and settling eight open Formal Conjectures problems, and variants of Erd\H{o}s~138~\citep{erdos138} and~337~\citep{erdos337}.
  At matched models, \hone{} surpasses Goedel-Architect~\citep{goedelarchitect2026} at lower estimated cost; on five open Erd\H{o}s problems, it Pareto-dominates full AlphaProof Nexus (APN)~\citep{alphaproofnexus2026} with the same LLM.
  \item \textbf{Longer attempts can be more efficient than repetition} (Section~\ref{sec:repetition}).
  With sequential test-time scaling, pass@1 traces a budget-dependent curve that can Pareto-dominate pass@$k$: on five Erd\H{o}s problems, extending \hone{} can cost less than repeating \hone{} or APN basic~\citep{alphaproofnexus2026}.
  \item \textbf{A simple harness still needs efficient interaction, reliable feedback, and sustained progress} (Section~\ref{sec:efficiency}).
  Among simple loops, harness design produces large differences in solve rates and cost. Controlled ablations and comparisons with AxProverBase~\citep{axprover}, Claude Code~\citep{claudecode2026}, and Humanfia~\citep{humanfia2026} show the value of a reliable and efficient interface to Lean with search tools, a loop that sustains repair and reflection, and efficient reasoning and cache reuse.
\end{itemize}

Our results show that a simple harness can match or outperform more complex systems, extending the bitter lesson to harness design in ATP and plausibly other structured environments.
Frontier models keep improving their reasoning and adaptation to execution feedback, and harnesses should help these capabilities scale through efficient interaction with the environment rather than impose workflows.
Future work should thus evaluate frontier models with efficient execution feedback before prescribing workflows, or risk a bitter lesson.

\section{Methods}

We first introduce the formal-proving task, benchmarks, and baselines, and explain how we evaluate test-time scaling. We then describe \hone{}'s tools and interaction loop, which let the model work efficiently and sustain progress as its inference budget grows.

\subsection{Experimental setup}

\paragraph{Formal proving.}
In Lean~\citep{moura2021lean4}, a formal statement specifies a mathematical claim and its assumptions; the agent must complete its proof in a source file.
Mathlib~\citep{mathlib4} is a shared library of formalized mathematics that provides definitions and previously proved results.
Compiling the file checks the proof and returns errors.
During proof development, \emph{open goals} list what remains to be proved.
Additionally, formal validation tools such as SafeVerify~\citep{safeverify} allow checking the final proof against the original statement, rejecting placeholders for unfinished steps and added assumptions.

\paragraph{Benchmarks.}
We evaluate on formal statements of increasing difficulty.
PutnamBench~\citep{tsoukalas2024putnambench} contains 672 problems from the past Putnam undergraduate competitions; for controlled comparisons, we use Putnam100, the same 100-problem subset used by AxProverBase~\citep{axprover}.
We also use the 12 problems of the 2025 Putnam competition, formalized both in PutnamBench and by Numina~\citep{numina2026leanagent}.
IMO~2026 contains the 6 problems of the 2026 International Mathematical Olympiad, the leading competition for high-school students, as formalized by Axiom~\citep{axiommath2026imo}.
FATE-X~\citep{fate2025} contains 100 algebra problems at PhD qualifying-exam level or beyond, and FormalQualBench~\citep{formalqualbench2026} 23 theorems at PhD qualifying-exam level.
Formal Conjectures~\citep{formalconjectures} collects formalized research conjectures; we evaluate on FC100-solved, a random sample of 100 problems with known informal solutions, and on open targets.
Appendix~\ref{app:benchmark-versions} records benchmark source revisions and Lean/Mathlib versions.
We omit older benchmarks such as miniF2F~\citep{zheng2021minif2f}, which modern models essentially saturate.
PutnamBench mainly tests efficiency, while FATE-X, FormalQualBench, and FC100-solved remain unsaturated at our scale.
Putnam~2025 and IMO~2026 are recent enough to be unseen by the models released before them (Appendix~\ref{app:chronology}).
Open conjectures, such as those in Formal Conjectures~\citep{formalconjectures}, test progress beyond problems with known solutions.

\paragraph{Models and baselines.}
We evaluate frontier models from OpenAI, Anthropic, and Google, alongside open-weight DeepSeek~V4~Flash~\citep{deepseekv4} and Leanstral~1.5~\citep{leanstral2026}.
We select baselines representing the most competitive results on our benchmarks as of August~27, 2026, alongside simple loops that help isolate the contribution of harness design.
All state-of-the-art (SOTA) claims use this comparison cutoff (Appendix~\ref{app:comparison-cutoff}).
To compare with human-designed workflows, we include systems that enforce planning, decomposition, or agent roles in code: Goedel-Architect~\citep{goedelarchitect2026}, MerLean~\citep{merlean2026prover}, and full AlphaProof Nexus~\citep{alphaproofnexus2026}.
OpenGauss~\citep{opengauss2026} instead prescribes a plan--review workflow to a general coding agent through instruction files (skills), while Numina~\citep{numina2026leanagent} gives a general coding agent optional tools that consult a natural-language (informal) prover and other models, and prescribes a subagent workflow for its hardest problem.
Comparisons with other simple loops test interaction efficiency and persistence; these include Humanfia~\citep{humanfia2026}, APN basic~\citep{alphaproofnexus2026} (AlphaProof Nexus's single-agent loop), and our same-model reruns of both AxProverBase~\citep{axprover} and Claude Code~\citep{claudecode2026}.
We also report the published scores of trained provers, such as Seed-Prover~1.5~\citep{chen2025seedprover15}, where they still define the SOTA.
Appendix~\ref{sec:related} discusses additional related work.

\paragraph{Measuring test-time scaling with cost per task.}
\label{sec:protocol}
Prior work very rarely compares systems at matched inference compute.
Earlier theorem-proving work focused on training better models, with evaluation emphasizing solve rates rather than inference cost.
As improvements increasingly come from inference-time search, repair, and orchestration, these reporting conventions have persisted even as an attempt has come to mean very different amounts of computation.
For example, on PutnamBench, one Goedel-Architect attempt permits up to 16 revisions of its proof plan~\citep{goedelarchitect2026}, while one AxProverBase attempt permits 50 agent iterations~\citep{axprover}.
Both papers report these as pass@1, but matching attempt counts do not match inference budgets; Appendix~\ref{app:putnam-followup} gives further examples.
Even with the same model, output tokens can cost up to $100\times$ as much as cached input
(Table~\ref{tab:model-token-prices}).
As inference-time methods drive more of the progress, accounting for their cost should become standard practice.
We compare inference cost, holding the model fixed to compare harnesses or the harness fixed to measure model progress.
Throughout the paper, dollar amounts for our experiments denote API-equivalent
costs computed from recorded token usage and the rates in
Table~\ref{tab:model-token-prices}.
At budget \$$B$ per problem, solve count includes problems with a verified proof within budget. Average cost includes all problems, including failures, with each problem's cost capped at \$$B$.

\subsection{\hone{}: a simple harness for efficient test-time scaling}
\label{sec:harness}

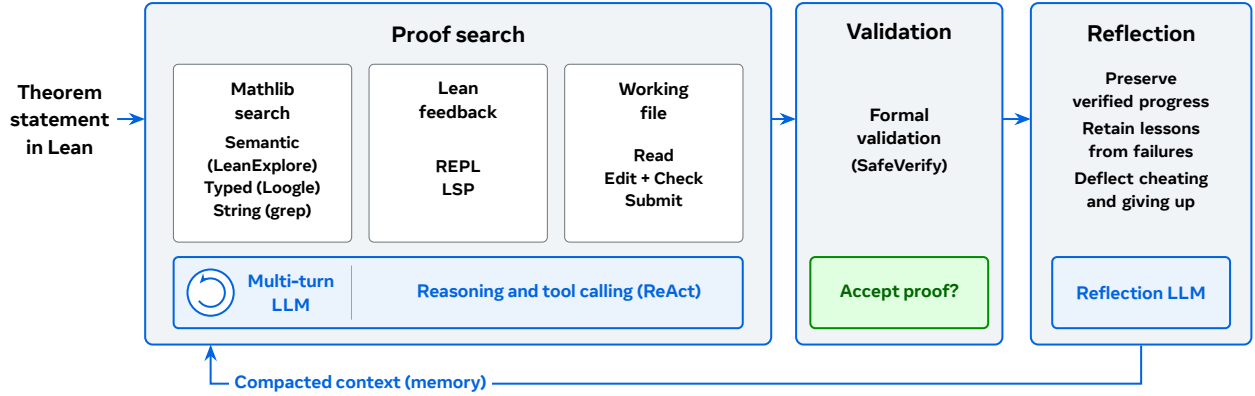
\begin{figure}[t]
\centering
\resizebox{\textwidth}{!}{%
\optimisticfont
\begin{tikzpicture}[
  >=Stealth,
  phase/.style={draw=metablue, thick, fill=metabg, rounded corners=3pt,
                minimum height=38mm, align=center, font=\small},
  tool/.style={draw=metafg!65, fill=white, rounded corners=2pt,
               minimum height=7mm, align=center, font=\footnotesize},
  contextcycle/.style={circle, draw=metablue, thick, fill=white,
                       minimum size=6.5mm, inner sep=0pt},
  contextstate/.style={draw=metablue, thick, fill=metablue!8, rounded corners=2pt,
                       minimum height=9mm, align=center},
  outcome/.style={draw=green!55!black, thick, fill=green!12, rounded corners=2pt,
                  minimum width=25mm, minimum height=10mm, inner sep=1pt,
                  align=center, font=\footnotesize\sffamily\bfseries, text=green!40!black},
  flow/.style={->, thick, draw=metablue},
  contextflow/.style={->, thick, draw=metablue}
]
  \path[draw=metablue,thick,fill=metabg,rounded corners=3pt]
    (-0.60,2.45) rectangle (8.20,-2.35);
  \coordinate (turnseast) at (8.20,0.82);
  \node[font=\fontsize{10}{12}\selectfont\sffamily\bfseries,anchor=north] at (3.80,2.45) [yshift=-2mm]
    {Proof search};
  \node[tool,minimum width=25mm,text width=23mm,minimum height=25mm] (search) at (1.05,0.34) {};
  \node[font=\footnotesize,align=center,anchor=north] at ([yshift=-1.5mm]search.north)
    {\textbf{Mathlib}\\\textbf{search}};
  \node[font=\fontsize{7.5}{9}\selectfont,align=center] at ([yshift=-3.5mm]search.center)
    {Semantic\\(LeanExplore)\\Typed (Loogle)\\String (grep)};
  \node[tool,minimum width=25mm,text width=23mm,minimum height=25mm] (lean) at (3.80,0.34) {};
  \node[font=\footnotesize,align=center,anchor=north] at ([yshift=-1.5mm]lean.north)
    {\textbf{Lean}\\\textbf{feedback}};
  \node[font=\fontsize{7.5}{9}\selectfont,align=center] at ([yshift=-3.5mm]lean.center)
    {REPL\\LSP};
  \node[tool,minimum width=25mm,text width=23mm,minimum height=25mm] (files) at (6.55,0.34) {};
  \node[font=\footnotesize,align=center,anchor=north] at ([yshift=-1.5mm]files.north)
    {\textbf{Working}\\\textbf{file}};
  \node[font=\fontsize{7.5}{9}\selectfont,align=center] at ([yshift=-3.5mm]files.center)
    {Read\\Edit + Check\\Submit};
  \node[contextstate,minimum width=80mm,minimum height=10mm] (livecontext) at (3.80,-1.62) {};
  \node[contextcycle] (llmloop) at (0.32,-1.62) {};
  \draw[metablue,thick,-{Stealth[length=1.4mm,width=1.1mm]}]
    ([shift=(215:2.0mm)]llmloop.center) arc[start angle=215,end angle=500,radius=2.0mm];
  \node[font=\footnotesize\sffamily\bfseries,text=metablue,align=center]
    at (1.45,-1.62) {Multi-turn\\LLM};
  \draw[metablue!55] (2.30,-2.00) -- (2.30,-1.24);
  \node[font=\footnotesize,text=metablue,align=center] at (5.22,-1.62)
    {Reasoning and tool calling (ReAct)};
  \coordinate (contextreturn) at (0.32,-2.12);
  \node[phase, minimum width=29mm,minimum height=48mm] (submit) at (10.00,0.05) {};
  \node[font=\fontsize{10}{12}\selectfont\sffamily\bfseries,anchor=north] at (submit.north) [yshift=-2mm]
    {Validation};
  \node[font=\fontsize{7.5}{9}\selectfont,align=center,text width=27mm] at (10.00,0.50)
    {{\footnotesize\textbf{Formal}}\\{\footnotesize\textbf{validation}}\\[2pt](SafeVerify)};
  \node[outcome] (accepted) at (10.00,-1.62)
    {Accept proof?};
  \node[phase, minimum width=31mm,minimum height=48mm] (reflect) at (13.40,0.05) {};
  \node[font=\fontsize{10}{12}\selectfont\sffamily\bfseries,anchor=north] at (reflect.north) [yshift=-2mm]
    {Reflection};
  \node[font=\fontsize{7.5}{9}\selectfont,align=center,text width=28mm] (reflectops) at (13.40,0.50)
    {Preserve\\verified progress\\[2pt]Retain lessons\\from failures\\[2pt]Deflect cheating\\and giving up};
  \node[contextstate,minimum width=25mm,minimum height=10mm,
        font=\footnotesize\sffamily\bfseries,text=metablue] (reflectllm) at (13.40,-1.62)
    {Reflection LLM};
  \draw[flow] (turnseast) -- (8.55,0.82);
  \draw[flow] (11.45,0.82) -- (11.85,0.82);
  \node[font=\small,align=center] (theorem) at (-1.80,0.82)
    {Theorem\\statement\\in Lean};
  \draw[flow] (theorem.east) -- (-0.60,0.82);
  \draw[contextflow] (reflect.south) -- (13.40,-2.90) -| (0.32,-2.35);
  \node[fill=white,inner xsep=2pt,inner ysep=0.5pt,
        font=\footnotesize\sffamily\bfseries,text=metablue,anchor=west]
    at (0.58,-2.90) {Compacted context (memory)};
\end{tikzpicture}
}
\caption{\textbf{\hone{}: a simple harness that scales.}
An LLM agent edits a Lean file through reasoning and tool calls (ReAct;~\citealp{yao2023react}) until it submits or reaches the $T$-turn limit, forcing submission. An external validator then checks the proof. On failure, a reflection LLM compacts the transcript and feedback for the next iteration. The loop provides Lean tools without prescribing a proof search structure. Search uses LeanExplore~\citep{leanexplore2025}, Loogle~\citep{loogle}, and grep; proof fragments run in the Lean read--eval--print loop (REPL)~\citep{leanrepl}; the language server (LSP) reports goals and definitions; SafeVerify~\citep{safeverify} performs validation.
}
\label{fig:harness}

\end{figure}

\hone{} keeps a model working on a Lean proof until an external validator accepts it or the budget runs out (Figure~\ref{fig:harness}).
In each \emph{iteration}, the model reasons and calls tools in a ReAct loop~\citep{yao2023react} for at most $T$ \emph{turns}, editing a persistent working file; every edit is compiled and returns errors and open goals.
The iteration ends when the model submits the file, which an external validator then checks against the original statement, also accepting checked disproofs where supported (Appendix~\ref{app:formal-validation}).
If the validator rejects the submission, a reflection LLM compacts the transcript and feedback into verified progress, lessons from failed attempts, and useful lemma names, and discourages giving up, as in Reflexion~\citep{shinn2023reflexion}, Self-Refine~\citep{madaan2023selfrefine}, and AxProverBase~\citep{axprover}; the model then continues from the current file with this summary.
The loop builds on APN basic~\citep{alphaproofnexus2026}, AxProverBase, and the \emph{Ralph} loop from software engineering~\citep{huntley2025ralph,macedo2026loops}.

\paragraph{A simple loop with an efficient interface.}
\hone{} imposes no decomposition, planning, or search structure.
Instead, it gives the model an efficient interface to Lean, with targeted edits rather than rewritten proofs, immediate compiler feedback, and a context that grows and is cached within an iteration.
Optional tools let the model search Mathlib, inspect definitions, and test proof fragments, and the model decides which to use and how to develop the proof.
Section~\ref{sec:efficiency} shows that these choices matter for both cost and success, and Appendix~\ref{app:harness} details tools and prompts.

\section{Simple test-time scaling delivers frontier results}
\label{sec:results}

We first show that frontier models in \hone{} reach state-of-the-art results across competition and research mathematics.
We then show that simple scaling often matches or outperforms specialized workflows at matched models and comparable inference costs, and that, even on open research problems, extending a single attempt can be more efficient than repetition.

\subsection{A frontier model in a simple harness achieves state-of-the-art results}
\label{sec:benchmarks}

\begin{figure}[t]
\centering
\begin{minipage}[t]{0.41\textwidth}
  \centering\includegraphics[width=\linewidth]{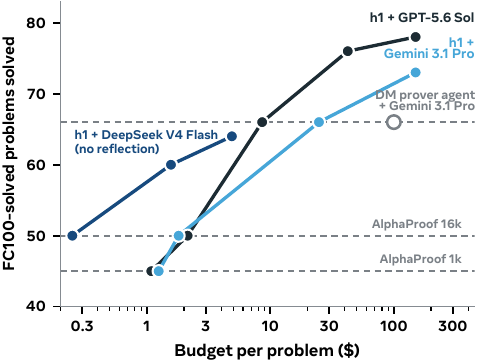}
\end{minipage}\hspace{0.06\textwidth}
\begin{minipage}[t]{0.41\textwidth}
  \centering\includegraphics[width=\linewidth]{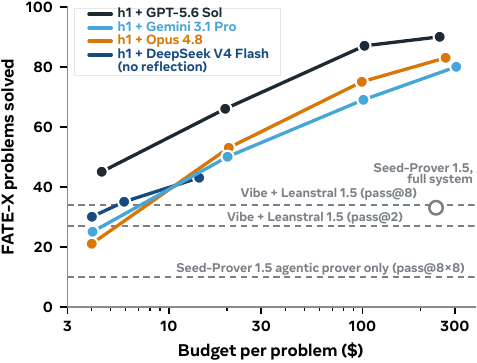}
\end{minipage}
\caption{\textbf{Generalist models reach frontier results through simple scaling.}
On FC100-solved (left) and FATE-X (right), increasing \hone{}'s inference budget takes generalist models beyond published benchmark scores.
On FC100-solved, \hone{} with Gemini~3.1~Pro matches the DeepMind (DM) prover agent's score at a quarter of its budget cap; with GPT-5.6~Sol, it solves 78, and 90 on FATE-X.
With DeepSeek~V4~Flash, \hone{} surpasses AlphaProof on FC100-solved and Leanstral~1.5 on FATE-X.
Baseline markers show the best published results as of August~27, 2026.}
\label{fig:frontiers}

\end{figure}

\paragraph{State-of-the-art results on established benchmarks.}
With GPT-5.6~Sol, \hone{} achieves state-of-the-art results: 90/100 on FATE-X, 78/100 on FC100-solved, 14/23 on FormalQualBench, and all six IMO 2026 problems (Figure~\ref{fig:frontiers}; Table~\ref{tab:complex}).
On PutnamBench, \hone{} with GPT-5.5 solves 670 of 672 problems.
All these results use a single run per problem, without repetition or parallelism.
Reaching state-of-the-art performance with a recent frontier model required only a larger inference budget, as the model keeps iterating, adjusting proofs and changing approaches without a prescribed workflow.
Across our runs, individual accepted attempts reached up to 1{,}123 rounds of reflection (compaction), 95 active hours, or 7{,}124 lines of proof (Appendix~\ref{app:search-horizon}).
An inexpensive open-weight model also surpasses specialized provers: DeepSeek~V4~Flash in \hone{} solves 43 FATE-X problems at a \$14.35 per-problem cap, exceeding Leanstral~1.5's published 34~\citep{leanstral2026}.
On FC100-solved, it matches the 50 solves~\citep{formalconjectures} of AlphaProof~\citep{alphaproof} below an estimated \$0.30 cap and reaches 64 at \$4.89 per problem, close to the DM prover agent's 66~\citep{dmproveragent2026}.
With Gemini~3.1~Pro, \hone{} also surpasses the DM prover agent on FC100-solved: it matches the agent's 66 solves at a \$25 rather than \$100 cap per problem, then reaches 73 at \$150~\citep{dmproveragent2026,formalconjectures}.\footnote{The DM prover agent's model is not named; we infer Gemini~3.1~Pro from its apparent connection to AlphaProof Nexus~\citep{alphaproofnexus2026}.}
In line with the bitter lesson, the gains frontier models draw from scaling training compute carry over to an unchanged simple loop: in \hone{}, GPT-5.6~Sol matches the DM prover agent's 66 FC100-solved solves below a \$9 cap, against \$25 for Gemini~3.1~Pro, and reaches 90 FATE-X solves at \$38 per problem, against 81 at \$113 (Figure~\ref{fig:frontiers}).

\paragraph{Solving open problems and exposing misformalizations.}
Beyond the solved subset of Formal Conjectures and benchmarks with possibly known informal solutions, we ran our system on 236 targets selected as open from Formal Conjectures (Appendix~\ref{app:open-solved}).
% TODO(Skander): state that each of these open targets also got a single run per problem (confirmed with Niket); the tweet thread says so.
It settled eight of them (Table~\ref{tab:open-solved}): variants of Erd\H{o}s~138 and~337, three OEIS conjectures (A108081, A211417, A022030), the existence of an AME$(9,10)$ state (Open Quantum Problem~35), and two conjectures of Written on the Wall~II.
We document concisely the newly proven theorems in Appendix~\ref{app:open-solved}, and our pull-request marking these as solved\footnote{\href{https://github.com/google-deepmind/formal-conjectures/pull/6008/}{https://github.com/google-deepmind/formal-conjectures/pull/6008/}}.
Equally important as finding these proofs was our system's ability to find misformalizations in Formal Conjectures: formal statements that do not express the intended conjecture.
Our system found several by producing short disproofs of the written statement, for instance for the Poincaré conjecture, whose formalization was missing a hypothesis.
Appendix~\ref{app:misformalizations} lists all misformalizations found, most of which have since been fixed upstream.

These results establish the capability of a simple harness, and we show in the next section that simple scaling often matches specialized workflows with the same model and comparable inference cost.

\begin{table}[t]
\centering
\caption{\textbf{Eight problems marked as open in Formal Conjectures proved by \hone{} with GPT-5.6 Sol.}}
\label{tab:open-solved}
\footnotesize
\setlength{\tabcolsep}{5pt}
\renewcommand{\arraystretch}{1.0}
\resizebox{0.95\linewidth}{!}{%
\begin{tabular}{@{}ll@{}}
\toprule
Problem & Statement proved \\
\midrule
Erd\H{o}s~138, variant & $W(k)/2^k\to\infty$ for the two-colour van der Waerden number \\
Erd\H{o}s~337, variant & $|(A+A)\cap[2N]|/|A\cap[N]|\to\infty$ for thin asymptotic bases (Ruzsa--Turj\'anyi) \\
OEIS A108081 & $|x_n| = a(n-1)$ for the inductively defined words \\
OEIS A211417 & $a(p^k)\equiv a(p^{k-1}) \pmod{p^{3k}}$ for primes $p\ge 5$ \\
OEIS A022030 & $a(n)=4a(n-1)-a(n-3)+a(n-4)$ for $n\ge 4$ \\
Open Quantum Problem~35 & an AME$(9,10)$ state exists \\
Written on the Wall~II, 100 & $\alpha(G)\le\lceil(\max_v l(v)+\tfrac12\|\deg G^c\|_2)/2\rceil$ for connected $G$ \\
Written on the Wall~II, 314 & connected triangle-free $G$ with induced paths of at most 4 vertices is well totally dominated \\
\bottomrule
\end{tabular}}

\end{table}

\subsection{The bitter lesson for specialized workflows}
\label{sec:complex}

We show that, at matched models, simple test-time scaling recovers much of the performance of specialized workflows, often at lower inference cost, and that repetition (pass@$k$) or a newer frontier model in the same loop usually closes any remaining gap (Figure~\ref{fig:teaser}; Table~\ref{tab:complex}).

\paragraph{Simple test-time scaling can match specialized systems at lower cost.}
On PutnamBench, one extended \hone{} attempt surpasses Goedel-Architect's blueprint-based search, which assigns separate agents to prove the lemmas in a dependency graph.
With the same DeepSeek~V4~Flash model, \hone{} first matches its pass@1 score at comparable cost, then reaches 598 solves against 597 at pass@4 with informal proof seeding, at an estimated \$1.08 versus \$1.47 per problem~\citep{goedelarchitect2026} (Figure~\ref{fig:teaser}, left).
Goedel-Architect's tool-integrated reasoning baseline reaches only 54.5\% at pass@32 on a 200-problem subset, compared with \hone{}'s 89.0\% at pass@1 on the full benchmark with the same model.
With Gemini~3.1~Pro alone, \hone{} also matches Goedel-Architect's July leaderboard result of 642 solves, which pairs Gemini~3.1~Pro with the cheaper DeepSeek~V4~Pro, at a similar cost (\$6.32 versus \$5.80 per problem).
The much stronger performance of a simple harness with the same model illustrates how a weak simple baseline can overstate the need for specialized workflows (Figure~\ref{fig:deepseek-reflection}).

\paragraph{A simple loop solves the problem that needed a hand-designed workflow.}
A single model in \hone{} also recovers the result of Numina-Lean-Agent, which reserved a specialized workflow for the hardest problem of Putnam 2025.
Numina runs Opus~4.5 in Claude Code, a general coding agent, with optional tools that call other models: an informal prover built as a Gemini~3~Pro generator--verifier loop, and a discussion partner~\citep{numina2026leanagent}.
For that problem, A5, the authors switch to a coordinator that delegates blueprint decomposition, sketching, and proving to subagents, seeded with a GPT-5.2 hint and a \$1{,}000 budget.
On the same statements, GPT-5.2 alone solves all twelve in \hone{}, including A5, in a single run per problem and without such a workflow, spending \$81.52
on A5 against Numina's \$1{,}000 budget (Table~\ref{tab:complex}).
Also note that with Opus~4.5 alone, \hone{} solves seven problems against four for Numina without its informal prover, consistent with our finding that Claude Code, on which Numina builds, scales less effectively than \hone{} with the same model (Section~\ref{sec:comparison}).

\paragraph{Repetition closes most remaining gaps.}
When a single \hone{} run falls short, rerunning the unsolved problems from a fresh context (residual pass@$k$) closes most remaining gaps.
Like \hone{} itself, repetition prescribes no workflow; it simply adds independent attempts, though often at a higher cost than extending a single run (Section~\ref{sec:repetition}).
As long as that cost is counted, including failed attempts, repetition therefore belongs to the simple baseline against which specialized workflows should be tested.
On Putnam 2025, \hone{} with Opus~4.7 solves eleven against MerLean's twelve; retrying the remaining problem recovers all twelve within six attempts.
On full PutnamBench, extra attempts on the last two problems let \hone{} with GPT-5.5 match Aleph Prover's~\citep{alephprover2026reports} final 672/672 coverage at an estimated \$17.86 versus about \$74 per problem, counting the cost of failed attempts (Appendix~\ref{app:putnam-followup}).
On FormalQualBench~\citep{formalqualbench2026}, OpenGauss~\citep{opengauss2026} instructs Claude Code to cycle through planning, proving, review, and replanning, generating several candidate proofs per open goal and extracting helper lemmas when a goal is stuck.
The best of four \hone{} runs with Opus~4.6 matches its eight solves under the same four-hour limit.
On these benchmarks, the only remaining exception after scaling and repetition is MerLean on FormalQualBench with Opus~4.7, which retains a one-problem advantage with or without the four-hour cutoff (Appendix~\ref{app:specialized-exceptions}).

\begin{table}[t]
\centering
\caption{\textbf{Simple scaling often matches specialized systems, while model progress closes further gaps.}
Row groups separate matched-model comparisons from those using other models. The 4-hour cap is FormalQualBench's official limit; $T50$/$T200$ allow 50/200 turns per iteration; \texttt{/goal} lets a small model judge completion. Table~\ref{tab:complex-with-time} adds measured times.}
\label{tab:complex}
\footnotesize
\setlength{\tabcolsep}{3.5pt}
\renewcommand{\arraystretch}{1.02}
\resizebox{0.94\textwidth}{!}{%
\begin{tabular}{@{}llccc@{}}
\toprule
Benchmark & System + model & Time cap & Solved & Cost \\
\midrule
\multirow{4}{*}{\shortstack[l]{Putnam 2025\\(Numina formalization)}}
  & Numina, no informal prover, Opus~4.5 & -- & 4/12 & -- \\
  & \textbf{\hone{} + Opus~4.5} & \textbf{--} & \textbf{7/12} & \$1{,}429 total \\
\cmidrule(lr){2-2}
  & \textbf{Numina, full system, Opus~4.5 + Gemini~3~Pro + GPT-5.2} & \textbf{--} & \textbf{12/12} & \$1{,}800 cap \\
  & \textbf{\hone{} + GPT-5.2} & \textbf{--} & \textbf{12/12} & \$565 total \\
\midrule
\multirow{6}{*}{\shortstack[l]{PutnamBench's\\2025 slice }}
  & \hone{} + Opus~4.7 & -- & 11/12 & \$25.97/prob. \\
  & \textbf{MerLean + Opus~4.7} & \textbf{--} & \textbf{12/12} & not reported \\
  & \textbf{\hone{} + Opus~4.7, pass@4} & \textbf{--} & \textbf{12/12} & \$105.82/prob. \\ %
  & \textbf{\hone{} + Opus~4.7, best@6} (cost excludes failures) & \textbf{--} & \textbf{12/12} & \$29.01/prob. \\
\cmidrule(lr){2-2}
    & \textbf{\hone{} + Opus~4.8 / Gemini~3.1~Pro / GPT-5.5} & \textbf{--} & \textbf{12/12} & \textbf{\$7.87 / \$7.12 / \$6.84/prob.} \\
\midrule
\multirow{4}{*}{IMO 2026}
  & AxiomProver & -- & 6/6 & not reported \\
  & \hone{} + Gemini~3.1~Pro & -- & 6/6 & \$54.37/prob. \\
  & Humanfia + GPT-5.6~Sol & -- & 6/6 & \$21.23/prob. \\
  & \textbf{\hone{} + GPT-5.6~Sol ($T50$ / $\boldsymbol{T200}$)} & \textbf{--} & \textbf{6/6} & \textbf{\$8.33 / \$8.07/prob.} \\
\midrule
\multirow{11}{*}{\shortstack[l]{FormalQualBench}}
  & Claude Code (math.inc run, Opus~4.6) & 4 h & 4/23 & \$23.16 \\
  & Claude Code (our rerun, Opus~4.6, \texttt{/goal}) & 4 h & 7/23 & \$17.28 \\
  & \textbf{OpenGauss (math.inc run, Opus~4.6)} & \textbf{4 h} & \textbf{8/23} & \textbf{\$24.93} \\
  & \textbf{\hone{} + Opus~4.6 (best seed of 4)} & \textbf{4 h} & \textbf{8/23} & \textbf{\$23.40} \\
\cmidrule(lr){2-2}
  & \hone{} + Opus~4.7 & 4 h & 8/23 & \$18.14 \\
  & MerLean + Opus~4.7 & 4 h & 9/23 & \$101.09 \\
  & \hone{} + Opus~4.7 & -- & 9/23 & \$43.92 \\
  & \textbf{MerLean + Opus~4.7} & \textbf{--} & \textbf{10/23} & \$118.35 \\
\cmidrule(lr){2-2}
  & \hone{} + GPT-5.6~Sol & 4 h & 11/23 & \$8.97 \\
  & \hone{} + Opus~4.8 & -- & 13/23 & \$49.46 \\
  & \textbf{\hone{} + GPT-5.6~Sol} & \textbf{--} & \textbf{14/23} & \textbf{\$26.38} \\
\bottomrule
\end{tabular}}

\end{table}

\paragraph{The next frontier model closes the rest.}
In some rare cases, such as on FormalQualBench with MerLean~\citep{merlean2026prover}, we find that repetition does not close the gap between \hone{} and the other methods.
However, even in this case, replacing Opus~4.7 with its direct successor, Opus~4.8, is enough for \hone{} to overtake MerLean, with 13 against 10 solves at \$49.46 rather than \$118.35 per solve, and with GPT-5.6~Sol, \hone{} also solves the one problem that only MerLean had proved (Table~\ref{tab:complex}).
MerLean would likely also improve with Opus~4.8 or GPT-5.6~Sol, but we expect this gap to keep shrinking as newer models internalize the strategies that workflows prescribe.
We see this as the bitter lesson at work, as newer models convert more training compute into capabilities that an unchanged simple loop directly benefits from.

\subsection{Persisting can beat repetition, even on open problems}
\label{sec:repetition}
When an attempt can keep going, pass@1 is no longer a single number but a curve that grows with budget: within one run, the model can abandon an approach and try another.
The question becomes whether spending more on one attempt beats spending the same on several.
AlphaProof Nexus (APN)~\citep{alphaproofnexus2026} found that its basic agent, a simple loop like \hone{}, can Pareto-dominate its full system, which evolves proof sketches with LLM raters, parallel provers, and AlphaProof, but only through repetition: at pass@$k$ for large enough $k$.
APN reports each pass@$k$, including pass@1, as a single point at a fixed budget, leaving open whether a single attempt given more budget can do what repetition does.

We evaluate \hone{} on the six open Erd\H{o}s problems that APN reports individually, using the same model, Gemini~3.1~Pro.
Following APN, we report mean cost conditional on success, charging each group its members' summed cost up to the first accepted proof.
This metric is undefined for a problem with no successful attempts, so Figure~\ref{fig:teaser} (middle) compares the five problems that all three systems solve at least once.
For \hone{}, we show how pass@1 changes as the budget grows, rather than reporting a single point.
As the budget grows, it matches APN basic's pass@1 at lower cost, then exceeds both APN basic's pass@3 (75.4\% at \$191) and full APN (76\% at \$317), reaching 82.5\% at \$121.
On these five problems, giving one \hone{} attempt more time achieves a higher solve rate at lower cost than three attempts of APN basic.
Full APN retains an advantage over both APN basic and \hone{} on the remaining problem, Erd\H{o}s~125, where all sixteen \hone{} attempts end in Lean timeouts (Appendix~\ref{app:erdos-aggregation}).

APN's conditional metric gives a first view of this depth--breadth trade-off, but omits spending on groups with no accepted proof (Appendix~\ref{app:erdos-aggregation}).
To account for these costs and separate compute allocation and isolate differences between harnesses, we also compare longer attempts with repetition within \hone{}, using mean spend across all successful and failed groups (Figure~\ref{fig:erdos-unconditional-aggregate}).
On the same five problems, pass@1 reaches 82.5\% at \$138 in mean spend including failures, whereas pass@3 needs \$202 to reach at least that rate.
Thus, the advantage of longer attempts at these budgets persists when failed searches are charged.
Repetition reaches higher success rates at larger budgets.
The right balance between depth (longer attempts) and breadth (more attempts) therefore depends on the budget and problem and is hard to set in advance.
The bitter lesson suggests leaving this choice to the model rather than to hand-set budgets.

\section{What a harness must provide for efficient test-time scaling}
\label{sec:efficiency}
\label{sec:comparison}
\label{sec:ablations}

\begin{figure}[t]
\centering
\begin{minipage}[t]{0.49\linewidth}
\vspace{0pt}%
\centering\includegraphics[width=\linewidth]{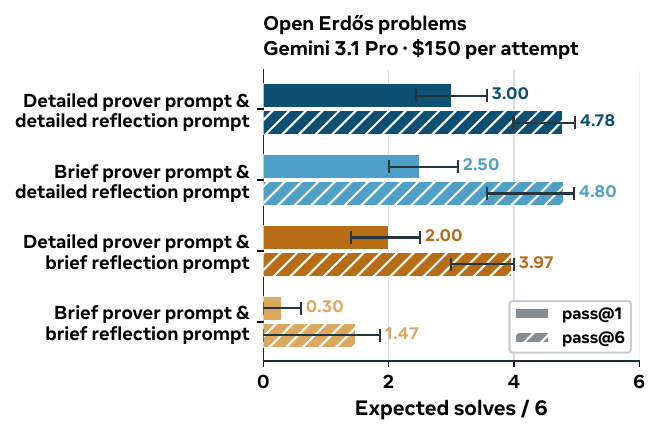}
\end{minipage}\hfill
\begin{minipage}[t]{0.49\linewidth}
\vspace{0pt}%
\centering\includegraphics[width=\linewidth]{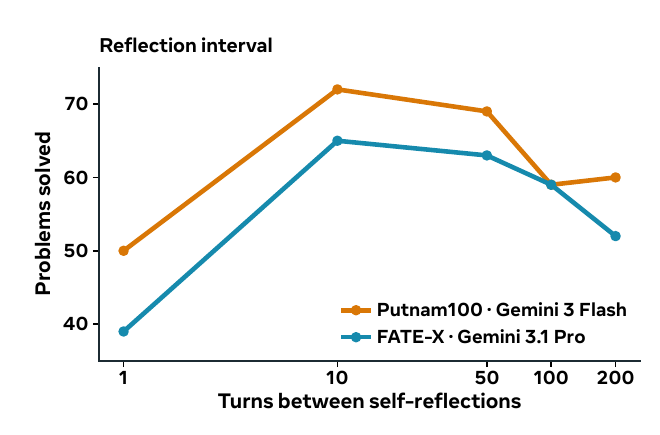}
\end{minipage}
\caption{\textbf{(Left) Detailed prompts sustain progress on open problems.} With Gemini~3.1~Pro on six open Erd\H{o}s problems and a \$150 cap per attempt, a single attempt with detailed prover and reflection prompts, which among other things deflect giving up and cheating, outperforms six attempts with brief ones.
\textbf{(Right) Moderate numbers of turns between reflections outperform both extremes.} Solves with Gemini~3~Flash on Putnam100 (\$20 cap) and Gemini~3.1~Pro on FATE-X (\$50 cap) as the number of turns between reflections varies.}
\label{fig:erdos-prompt-main}
\label{fig:ablation-summary}

\end{figure}

Leaving the proof workflow to the model does not make the harness irrelevant: with the same model, simple loops differ several-fold in solve rate and cost.
We explain these differences, and what makes \hone{} a state-of-the-art harness and an effective Lean baseline, by pairing comparisons with other simple loops with ablations on \hone{} that vary its tools, prompts, reflection, turns between reflections, and reasoning effort.
The resulting lessons concern the interface, persistence, and context management, and should transfer beyond the automated theorem proving domain, to agents that work against a verifier in rich environments allowing structured development and retrieval, such as software verification and coding in large libraries and codebases.

\afterpage{%
\begin{figure}[t]
\centering
\begin{minipage}[c]{0.49\linewidth}
\centering\includegraphics[width=0.8157\linewidth]{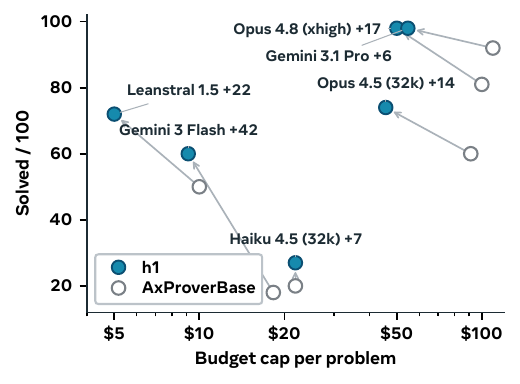}
\end{minipage}\hfill
\begin{minipage}[c]{0.49\linewidth}
\centering\includegraphics[width=0.9\linewidth]{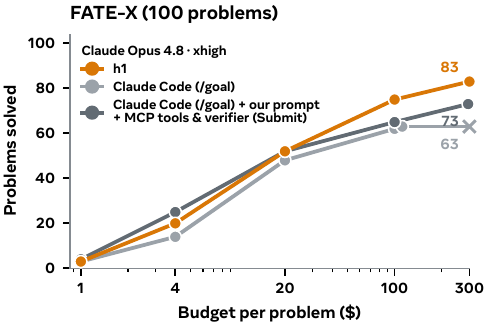}
\end{minipage}
\caption{\textbf{(Left) \hone{} Pareto-dominates AxProverBase on Putnam100.} For each of six model settings, the arrow goes from AxProverBase to \hone{}, labelled with the gain in solves at a comparable or lower budget cap.
\textbf{(Right) Claude Code improves with our Lean tools but still falls short of \hone{} on FATE-X.} With Opus~4.8, replacing Claude Code's built-in shell and file tools with our prompt and Lean tools through the Model Context Protocol (MCP) raises its solves from 63 to 73, while \hone{}, which also enforces validation, repair, and reflection, reaches 83 (Putnam100: Figure~\ref{fig:claude-code-putnam}).}
\label{fig:putnam100-pair}
\label{fig:claude-code-cap-curves}

\end{figure}
}

\paragraph{An efficient, reliable interface lets the model act in small steps and look things up.}
\label{sec:axprover}
\label{sec:tools}
In \hone{}, the model edits a persistent working file over many turns, sees compiler feedback after every edit, and can search the library or test a proof fragment at any point.
AxProverBase~\citep{axprover} instead makes one round of tool calls, then regenerates a complete proof after each round of feedback.
With the same Gemini~3~Flash model and a \$20 budget, it solves 18 Putnam100 problems against \hone{}'s 69, and trails in all six settings we compare (Figure~\ref{fig:putnam100-pair}, left).
A reliable interface also removes avoidable failures by construction: 42.2\% of AxProverBase's failed compilations stop at an invalid library import, which cannot occur in \hone{} because its file already imports all of Mathlib (Figure~\ref{fig:comparison-extra}).
Likewise, \hone{} retries transient API failures and restores the file after a Lean timeout, whereas some Claude Code runs failed through API errors and repeated timeouts even after restarts (Appendix~\ref{app:claude-code-mechanisms}).
Library search, which AxProverBase found the least useful of its components, matters much more when the model can act on its results over many turns: without search tools, the REPL, or the language server, \hone{} solves 49 Putnam100 and 49 FATE-X problems, against 69 and 63 with all tools (Figure~\ref{fig:tools-ablation}).
Any single search tool or the REPL recovers most of this gain, and removing any one tool from the full set costs at most four solves: what matters is the ability to look up and check, not a particular tool.
Other simple loops with less efficient interfaces reach \hone{}'s solve rates only at higher cost.
APN basic~\citep{alphaproofnexus2026} lacks dedicated search tools; on the five-problem Erd\H{o}s subset, extending a single \hone{} attempt can be more efficient than repeating APN basic (Section~\ref{sec:repetition}).
Humanfia~\citep{humanfia2026} checks proofs only through shell commands and restarts its coding agent in a fresh session each round, carrying over only the proof file and reviewer feedback; with GPT-5.5, it spends five times more than \hone{} for the same 670 PutnamBench solves.
A harness should thus support the model by offering a persistent workspace edited in small steps, immediate feedback, ways to retrieve and check what the model needs, and robust handling of predictable failures, leaving their use to the model.

\paragraph{The loop, not the model, decides when to stop.}
\label{sec:claude-code}
\label{sec:prompt-reflection}
Test-time compute helps only if the harness turns a larger budget into continued work.
With Opus~4.8, native Claude Code~\citep{claudecode2026} solves 63 FATE-X problems; giving it our prompt and our Lean tools through the Model Context Protocol (MCP), in place of its built-in shell and file tools, raises this to 73, while \hone{} reaches 83 (Figure~\ref{fig:claude-code-cap-curves}, right; 81, 95, and 99 on Putnam100).
Access to verification helps, but the model can still stop without calling the validator or give up after a rejection; in \hone{}, a rejection starts another iteration with the working file, the feedback, and a reflection summary (Appendix~\ref{app:claude-code-mechanisms}).
Within \hone{}, the system prompts matter most on open problems. Unlike one-line prompts, our detailed prompts describe the tools, set rules for statement integrity and disproofs, and tell the prover not to give up because a problem is open and the reflection LLM not to endorse such a decision (Appendix~\ref{app:prompts}). On six Erd\H{o}s problems, a single attempt with them outperforms six attempts with one-line prompts at the same \$150 cap per attempt (Figure~\ref{fig:erdos-prompt-main}, left).
Reflection, which carries verified progress and lessons into the next iteration, raises \hone{} with DeepSeek~V4~Flash from 554 to 598 PutnamBench solves, consistent with AxProverBase's findings (Figure~\ref{fig:deepseek-reflection}).
None of these tells the model how to prove: the harness decides whether the model keeps going, not how.

\paragraph{Reuse context, reset it before it degrades, and think more per turn.}
\label{sec:cadence}
\label{sec:cost-breakdown}
Each turn in a continuing context rereads it from cache, typically ten times cheaper than uncached input and a hundred times cheaper than output, whereas each restart discards it.
AxProverBase sits at one extreme: output accounts for 89\% of its \$17.39 per problem, against 34\% of \hone{}'s \$9.00, and it reuses at most 0.5\% of its input from cache, against 74--94\% for \hone{} (Figure~\ref{fig:cost-breakdown}).
Varying the number of turns between reflections in \hone{} exposes the same trade-off (Figure~\ref{fig:ablation-summary}, right).
With one turn, \hone{} solves 50 Putnam100 and 39 FATE-X problems, against 72 and 65 with ten turns, and pays roughly twice as much per solve.
With 200 turns instead of one, the cached share of input rises from 0.2\% to 86\%, yet solves drop to 60 and 52 (Figure~\ref{fig:effort-econ}): caching makes long contexts cheap but not productive, consistent with context rot~\citep{hong2025context}.
Thinking more per turn is a better use of budget.
AxProverBase found that raising the thinking budget from 10k to 32k tokens gains as much as doubling the number of iterations, at lower cost.
In \hone{}, higher reasoning effort solves more at lower total cost: with Gemini~3.1~Pro, it solves 89 Putnam100 problems at \$20.63 each with low effort and 99 at \$5.34 with high effort, needing far fewer compilations (Figure~\ref{fig:reasoning-effort}).
In short, a harness for long-horizon agents should let the model work in one context long enough to reuse it, compact that context before it degrades, and favor reasoning over repeated failed attempts.

\section{Conclusion, limitations, and discussion}
\label{sec:discussion}
\label{sec:conclusion}
\label{sec:limitations}
\label{sec:broader-implications}

\paragraph{Conclusion.}
We asked whether harnesses for frontier models should prescribe hand-designed workflows or give the model a simple loop and a larger budget.
With \hone{}, a simple loop with an efficient interface to Lean, test-time compute alone reaches state-of-the-art results and matches or surpasses hand-designed workflows at the same model and comparable cost. We find that a single longer attempt in this loop can even outperform repetition on open problems.
Getting the details of the harness right still matters: its interface, persistence, and context management determine how efficiently compute turns into progress.
Before designing a workflow, one should thus carefully try a frontier model in a simple loop at a realistic budget.

\paragraph{Limitations and discussion.}
We acknowledge that most frontier models are black boxes, and that we cannot measure the extent to which benchmark exposure during training could narrow the gap between specialized workflows and a simple loop, but where possible we used models that predate the problems or their solutions, such as Opus~4.5 and GPT-5.2 for Putnam~2025 and GPT-5.6~Sol for IMO~2026 (Appendix~\ref{app:chronology}).
Across the 707 benchmark problems on which we compare the simple loop with a specialized workflow, we find only two exceptions where the specialized harness is able to solve a problem we are not\footnote{Full APN solves Erd\H{o}s~125, where all sixteen \hone{} attempts end in Lean timeouts, and MerLean with Opus~4.7 proves \texttt{BurnsidePrimeDegree}, which \hone{} with Opus~4.8 does not prove in our run, although \hone{} with GPT-5.6~Sol does (Appendix~\ref{app:specialized-exceptions}).}.

Regarding scale, \hone{} works on a single file, whereas formalizations that exceed a context window, such as those of Fermat's Last Theorem~\citep{anthropic2026flt} or entire textbooks~\citep{rammal2026autoformbot}, require dependency management across files, and codebase-level validation, as well as parallel agents that share intermediate results to complete the task in reasonable time: in OpenAI's Navier--Stokes resolution, about 10{,}000 agents exchanged millions of messages over 88 hours~\citep{openai2026navierstokes}.
Our tools also cover only Lean, not numerical computation, simulation, visualization, or web search.
What makes a simple loop efficient and effective at that scale is the natural next question.
Following the bitter lesson, parallel agents, codebase management, and richer tools should be added without prescribing their use, for instance by letting agents communicate freely rather than through a fixed hierarchy.

\hone{}'s interface and tools are themselves hand-designed, which the bitter lesson might discourage.
Yet human design and leveraging computation need not run counter to each other~\citep{sutton2019bitterlesson}; the question is where each helps most first.
In formal mathematics, spending compute on the reasoning of frontier models, and using human effort on the interface currently works best.
Further ahead, a model could build and deploy its own harness for a problem and become more efficient than the human-designed interfaces.

\clearpage
\section*{Acknowledgements}

JK and NP thank the Simons Foundation for support through the Collaborative Grant “The Physics of Learning and Neural Computation” as well as support by the NSF through NRT Award 1922658. 
The authors would additionally like to thank Vivien Cabannes, Charles Arnal, and Amaury Hayat for useful discussions and feedback.

\bibliographystyle{iclr2027_conference}
\bibliography{paper}

\clearpage
\appendix
\setcounter{topnumber}{2}
\setcounter{totalnumber}{3}
\raggedbottom
\addtocontents{toc}{\protect\setcounter{tocdepth}{2}}

\begingroup
\renewcommand{\contentsname}{Appendix contents}
\tableofcontents
\endgroup
\clearpage

\section{Related Work}
\label{sec:related}

\paragraph{Trained provers.}
Neural provers began with tactic prediction inside proof search~\citep{polu2020gptf,lample2022htps,polu2022curriculum,yang2023leandojo,kumarappan2024leanagent,song2024leancopilot,george2025leanprogress,lin2024leanstar,xin2025bfsprover}, reaching olympiad level with AlphaProof's reinforcement learning and search~\citep{alphaproof}.
Later models generate whole proofs~\citep{first2023baldur,xin2024dsproverv15,ren2025dsproverv2,kimina2025prover,dong2025stp,goedelprover2025,goedelproverv2_2025} or are trained for agentic, tool-integrated proving~\citep{chen2025seedprover15,wang2026longcatprover,ma2026oprover}.
These systems contribute trained models rather than a harness around a given model, so we compare against their reported scores where available (Section~\ref{sec:benchmarks}).

\paragraph{Harnesses around generalist models.}
Following informal-to-formal sketching~\citep{jiang2023dsp} and early agentic loops~\citep{thakur2023copra,ospanov2025apollo}, recent systems add proof-specific structure around generalist models: auxiliary lemmas and informal reasoners~\citep{baba2025proveragent,liu2026dap}, multi-agent proving~\citep{breen2025axprover}, recursive decomposition~\citep{deltaprover2025,hilbert2025,merlean2026prover}, blueprints~\citep{goedelarchitect2026}, optional informal-prover and multi-model tools~\citep{numina2026leanagent}, and evolutionary search over proof sketches in full APN~\citep{alphaproofnexus2026}; Seed-Prover~\citep{seedprover2025} and Aristotle~\citep{harmonic2025aristotle} build such structure around their own trained provers.
We compare with several of these using the same model (Section~\ref{sec:complex}).
\citet{rognvaldsson2026costquality} reduce the compute such workflows spend on failed attempts with a learned controller; we instead remove the workflow.
Among simple loops, we compare with AxProverBase~\citep{axprover}, Claude Code~\citep{claudecode2026,formalqualbench2026}, Humanfia's loop with reviewer feedback~\citep{humanfia2026}, and APN basic~\citep{alphaproofnexus2026} (Section~\ref{sec:comparison}).
AxProverBase demonstrates the strength of minimal agents and self-reflection for retaining lessons~\citep{axprover}.
We extend this evidence to open problems and compare harnesses using the same models and accounting for inference cost.
These agents use Lean interfaces including the REPL~\citep{leanrepl}, Pantograph~\citep{aniva2024pantograph}, the Kimina server~\citep{kimina2025server}, and lean-lsp-mcp~\citep{dressler2025leanlspmcp}.

\paragraph{Autoformalization systems.}
Aleph~\citep{alephprover2026}, AxiomProver~\citep{axiommath2025axiom,axiommath2026imo}, and OpenGauss~\citep{opengauss2026} target formalization more broadly, as do efforts on major results~\citep{anthropic2026flt}, corpora~\citep{rammal2026autoformbot}, and research mathematics~\citep{zhang2026leanmarathon,moakhar2026theo}.
OpenGauss prescribes its planning and review workflow through skill instructions, whereas Goedel-Architect enforces its blueprint search in code.
We compare with Aleph, AxiomProver, and OpenGauss only on proof completion for fixed statements, which evaluates part of what they do; our results say nothing about their value in open-ended formalization (Section~\ref{sec:limitations}).

\paragraph{Test-time scaling and simple loops.}
Inference compute scales through repeated sampling~\citep{brown2024monkeys,snell2024scaling} and longer reasoning~\citep{openai2024o1,deepseekr1}; \hone{} extends s1's budget forcing~\citep{muennighoff2025s1} from one chain of thought to an agent's iterations.
For agents, test-time scaling has been studied in general~\citep{zhu2025agentscaling} and in coding~\citep{kim2026agenticcoding,ding2026swereplay}; \citet{li2026generalagentbench} find sequential scaling limited by context and parallel scaling by verification, the two limitations addressed by \hone{}'s reflection and verifier.
Rather than hand-crafting test-time scaling strategies, \citet{zheng2026autotts} have an agent discover them.
\hone{} combines ReAct~\citep{yao2023react} with self-reflection~\citep{shinn2023reflexion,madaan2023selfrefine}, and enforces external validation, which LLMs need to self-correct~\citep{huang2024selfcorrect,kambhampati2024llmmodulo}.
In software engineering~\citep{jimenez2024swebench}, interfaces matter~\citep{yang2024sweagent}, simple pipelines and minimal agents compete with complex ones~\citep{xia2024agentless,minisweagent2025}, and the Ralph loop runs an agent until verification succeeds~\citep{huntley2025ralph,macedo2026loops}.
Following \citet{kapoor2024agentsmatter}, we evaluate accuracy jointly with cost (Section~\ref{sec:protocol}).

\paragraph{Benchmarks and validation.}
Beyond earlier benchmarks~\citep{zheng2021minif2f,ying2024leanworkbook,yu2025formalmath,ravi2026formalproofbench,letson2026sorrydb}, we evaluate on PutnamBench~\citep{tsoukalas2024putnambench}, FATE~\citep{fate2025}, FormalQualBench~\citep{formalqualbench2026}, and Formal Conjectures~\citep{formalconjectures}, accepting proofs through a formal validator~\citep{safeverify,comparator,axle2026verifyproof} since compilation alone need not prove the intended statement.

\clearpage

\clearpage

\section{Tools, Prompts, and Error Handling in \hone{}}
\label{app:harness}

This appendix specifies the tools, prompts, and error handling used by \hone{}
(Section~\ref{sec:harness}): tool interfaces (Table~\ref{tab:tools}), prompt variants
(Table~\ref{tab:prompt-variants}), reflection safeguards
(Table~\ref{tab:reflection-failures}), and Lean failure handling
(Table~\ref{tab:lean-recovery}). Formal validation is described in
Appendix~\ref{app:formal-validation}.

\subsection{Tools}
\label{app:tools}

\begin{table}[H]
\centering
\caption{\textbf{Tools available to \hone{}.} File tools let the multi-turn LLM inspect, edit,
and submit its proof. Search tools retrieve Mathlib declarations; the REPL checks
standalone Lean snippets; language-server tools expose goals, diagnostics, and
symbol information.}
\label{tab:tools}
\footnotesize
\setlength{\tabcolsep}{4pt}
\renewcommand{\arraystretch}{1.0}
\begin{tabularx}{\textwidth}{@{}>{\raggedright\arraybackslash}p{0.48\textwidth}X@{}}
\toprule
Tool and arguments & Behavior \\
\midrule
\texttt{view\_working\_file}\newline {\ttfamily\scriptsize (offset, limit\,|\,None)} & Line-numbered, paginated view of the mutable file; 40,000-byte view limit. \\
\texttt{view\_original\_file}\newline {\ttfamily\scriptsize (offset, limit\,|\,None)} & The immutable statement file with its initial compile feedback; 40,000-byte view limit. \\
\texttt{edit\_working\_file}\newline {\ttfamily\scriptsize (old\_string, new\_string, replace\_all)} & Whitespace-tolerant replacement; unique unless \texttt{replace\_all}. Compiles after editing and returns errors and open goals (32,000 characters). Invalid import edits are rolled back. \\
\texttt{undo\_edit\_working\_file}\newline {\ttfamily\scriptsize ()} & Revert the last edit and recompile; compiler feedback: 32,000 characters. \\
\texttt{submit\_working\_file}\newline {\ttfamily\scriptsize (updated\_file, reasoning)} & Check for errors and unfinished goals, then compile and validate with SafeVerify in an isolated environment; ends the iteration. \\
\midrule
\texttt{lean\_leanexplore}\newline {\ttfamily\scriptsize (query)} & Natural-language search over Mathlib declarations (LeanExplore; Qwen3 embedding and reranker); 6 results with name, module, signature, description. \\
\texttt{lean\_loogle}\newline {\ttfamily\scriptsize (query)} & Search by name or type pattern (Loogle); 6 results. \\
\texttt{grep\_mathlib}\newline {\ttfamily\scriptsize (pattern, glob\,|\,None, context, fixed)} & ripgrep over the Mathlib sources with optional glob and context; at most 50 matching lines. \\
\texttt{lean\_repl\_run\_scratch\_code}\newline {\ttfamily\scriptsize (code)} & Elaborate a self-contained fragment in the same import environment; does not see or change the working file. \\
\texttt{lean\_goal}, \texttt{lean\_term\_goal}\newline {\ttfamily\scriptsize (line, column, file\_path\,|\,None)} & Show local hypotheses and remaining proof goals (\texttt{lean\_goal}), or the expected type of a term (\texttt{lean\_term\_goal}), at the selected position. \\
\texttt{lean\_diagnostic\_messages}\newline {\ttfamily\scriptsize (file\_path\,|\,None)} & List errors, warnings, and informational messages with their positions in the working file. \\
\texttt{lean\_hover\_info}\newline {\ttfamily\scriptsize (line, symbol\,|\,None, column\,|\,None, file\_path\,|\,None)} & Show an identifier's type signature and documentation, including definitions from Mathlib or the problem itself. \\
\texttt{lean\_declaration\_file}\newline {\ttfamily\scriptsize (line, symbol\,|\,None, column\,|\,None, file\_path\,|\,None)} & Locate an identifier's definition and display a source excerpt, exposing its implementation or structure fields. \\
\texttt{lean\_references}\newline {\ttfamily\scriptsize (line, symbol\,|\,None, column\,|\,None, file\_path\,|\,None)} & List source locations where an identifier is used, including its declaration; return up to 40 references. \\
\texttt{lean\_file\_outline}\newline {\ttfamily\scriptsize (file\_path\,|\,None)} & List the working file's declarations and available type signatures, organized by their nesting. \\
\midrule
\multicolumn{2}{@{}l@{}}{Generic replies: 6,000 characters. Language-server replies: 4,000 characters.} \\
\bottomrule
\end{tabularx}
\end{table}

\subsection{Prompts for proof search and reflection}
\label{app:prompts}

\hone{} uses two LLM roles (Figure~\ref{fig:harness}). The multi-turn LLM performs
proof search by editing the working file and calling tools. After validation
fails, the reflection LLM summarizes the attempt and verifier feedback into
compacted context for the next iteration. The multi-turn LLM receives this
context alongside the original statement, current working file, and verifier feedback.
Our Detailed prompt is adapted from the prompts of AxProverBase~\citep{axprover}
and APN basic~\citep{alphaproofnexus2026}.
Table~\ref{tab:prompt-variants} compares the Brief and Detailed system prompts
for the multi-turn LLM and the reflection LLM. Each LLM has these two prompt
versions. The Detailed prompts are reproduced in the labeled boxes below.

\begin{table}[H]
\centering
\caption{\textbf{Brief and Detailed system prompts.} Each LLM independently uses either version, giving four prompt combinations. Detailed prompts include general advice and examples without imposing a fixed search structure or decomposition procedure. Tool descriptions are included only when the corresponding tool is available.}
\label{tab:prompt-variants}
\footnotesize
\setlength{\tabcolsep}{4pt}
\renewcommand{\arraystretch}{1.12}
\begin{tabularx}{\textwidth}{@{}p{0.12\textwidth}XX@{}}
\toprule
Prompt version & Multi-turn LLM & Reflection LLM \\
\midrule
Brief & \textsf{``Prove the Lean file in the user message. Use the available tools and submit your completed working file.''} & \textsf{``Summarize the prover's progress for its next attempt. Be concise and return only the summary.''} \\
Detailed & Describes tool use and iterative state; specifies statement integrity, proof-validity and disproof requirements, and persistence. Includes general advice and examples. & Instructs faithful context compaction: preserve verified progress and useful lemma names, attribute unverified claims to the multi-turn LLM, and treat incomplete submissions as checkpoints. \\
\bottomrule
\end{tabularx}
\end{table}

\begin{table}[H]
\centering
\caption{\textbf{Observed failures and corresponding instructions in the Detailed prompts.}}
\label{tab:reflection-failures}
\footnotesize
\setlength{\tabcolsep}{4pt}
\renewcommand{\arraystretch}{1.12}
\begin{tabularx}{\textwidth}{@{}p{0.33\textwidth}X@{}}
\toprule
Observed failure & Prompt response \\
\midrule
The multi-turn LLM uses the Lean REPL like a shell to hunt through the sandbox filesystem for a stored solution & The sandbox contains no solutions. The prompt explains that searching for stored answers or guessing proof names is wasted effort and directs the model to search for supporting lemmas instead. \\
The reflection LLM treats an incomplete checkpoint as a final failure & The submission tool asks the multi-turn LLM to explain why it is submitting. This reason is passed to the reflection LLM, whose prompt distinguishes a claimed complete proof from a checkpoint submitted when a limit is reached and asks it to preserve partial progress and remaining goals. \\
The reflection LLM turns an unverified claim into a fact & Attribute mathematical beliefs to the multi-turn LLM; state file contents and verifier feedback directly. \\
The reflection LLM reinforces a decision to give up & Retain the concrete unresolved obstacle without repeating claims that the problem is impossible or too difficult. \\
\bottomrule
\end{tabularx}
\end{table}

\paragraph{Detailed prompt text.}
The boxes reproduce both LLMs' detailed system prompts
with all tools enabled, adapting only Markdown formatting for print. Tool-specific sections are omitted when those tools are unavailable.
Table~\ref{tab:prompt-variants} quotes both brief prompts; the split ablations exchange them.
User messages insert the original statement, initial Lean feedback, current proof,
latest attempt, and compacted context; the reflection LLM also receives the transcript, except with \texttt{no\_transcript}.

\begin{promptquote}{Multi-turn LLM: detailed system prompt}
You are a world-class research mathematician and Lean 4 expert, working in an ITERATIVE proof development process.
\par
\par\smallskip\noindent\textbf{Task}\par\nopagebreak
\par
Your task is to propose the next iteration of a Lean 4 proof. The execution feedback (successes, errors, remaining goals) informs the next iteration.
Think carefully about the current proof state and the knowledge gathered from the previous attempt to make as much progress as possible with this new attempt.
\par
\par\smallskip\noindent\textbf{Iterative state}\par\nopagebreak
\par
You work in \textbf{iterations}. A new iteration begins with the current proof state, a SUMMARY of the previous iteration, and its verifier feedback. An iteration ends when the proof is submitted or its turn budget is exhausted. Repeat until the proof is verified.
\par
Each iteration's user message shows \texttt{<file>} --- the ORIGINAL problem statement, for reference; it never changes --- and, from the second iteration on, \texttt{<code>} --- your working file as you left it last iteration. Your tools edit the working file (the \texttt{<code>} state); \texttt{<file>} is only the canonical goal to check yourself against, so if your working statement ever drifts from it, restore it.
\par
\par\smallskip\noindent\textbf{Tool: working-file editing}\par\nopagebreak
\par
You do NOT paste the proof into your chat reply --- build the proof by EDITING the working file with your tools, and the file you leave behind IS your submission. Every successful edit automatically compiles that file.
\par
\par\smallskip\noindent\textbf{Tool: scratch Lean execution}\par\nopagebreak
\par
Use \texttt{lean\_repl\_run\_scratch\_code} only for self-contained scratch experiments; it does not check or change the working file. Scratch code runs in the same warm import environment defined by the working file's current import header, and imports written in scratch code are ignored.
\par
\par\smallskip\noindent\textbf{Tool: Mathlib source search}\par\nopagebreak
\par
Use \texttt{grep\_mathlib} to search Mathlib source when you need exact declarations, nearby lemmas, or usage examples.
\par
\par\smallskip\noindent\textbf{Tools: declaration lookup}\par\nopagebreak
\par
The environment may include definitions beyond Mathlib --- the problem can rely on additional imported libraries, so an unfamiliar identifier is not necessarily a mistake. Look up any identifier you don't recognize with \texttt{lean\_hover\_info} / \texttt{lean\_declaration\_file} --- they resolve the definition, signature, docstring, or source of the symbol, INCLUDING the non-Mathlib ones.
\par
\par\smallskip\noindent\textbf{Tools: lemma search}\par\nopagebreak
\par
Use \texttt{lean\_leanexplore} (natural-language) and \texttt{lean\_loogle} (name / type-pattern) to find the Mathlib lemmas your proof will need. Use the lookup and search tools to find REAL definitions and REAL supporting lemmas, then prove the target yourself.
\par
\par\smallskip\noindent\textbf{Tool: final submission}\par\nopagebreak
\par
When you are confident the proof is COMPLETE and free of \texttt{sorry}, call \texttt{submit\_working\_file}. It runs the strong kernel-level verifier (SafeVerify), and if it accepts you are DONE. A submission ends the iteration.
\par
\par\smallskip\noindent\textbf{Proof development}\par\nopagebreak
\par
\par\noindent\textbullet\enspace \textbf{Incomplete proofs can help}: You may use incomplete proofs to strategically explore proof structure and understand what goals need to be proven. However, they are not a valid final result
\par\noindent\textbullet\enspace \textbf{How `sorry` is handled}: after every successful edit, the harness compiles the working file and reports the unsolved goal at each remaining \texttt{sorry} (which stays in place).
\par\noindent\textbullet\enspace \textbf{Execution feedback is information}: Compilation errors and goal states help you understand what to try next
\par\noindent\textbullet\enspace \textbf{Experience is valuable}: We keep track of the main lessons learned across all previous attempts. Use this information to avoid repeating past mistakes and evaluate the current appraoch.
\par\noindent\textbullet\enspace \textbf{Make steady progress}: Each iteration and tool use counts! Therefore, you should aim to fix existing errors, complete missing parts of the proof, or even explore a new approach if the current one is clearly not working.
\par\noindent\textbullet\enspace \textbf{First attempts}: Try a promising approach and see what feedback you get
\par\noindent\textbullet\enspace \textbf{Later iterations}: Analyze the feedback from the previous build and address the issues directly
\par\noindent\textbullet\enspace Each tool reply shows how many turns remain: explore early (leaving \texttt{sorry}s to inspect goals is fine), consolidate into a complete, \texttt{sorry}-free proof as it runs low.
\par
\par\smallskip\noindent\textbf{Quality standards}\par\nopagebreak
\par
\par\noindent\textbullet\enspace Use appropriate Lean 4 tactics and syntax.
\par\noindent\textbullet\enspace Reference Mathlib lemmas when applicable.
\par\noindent\textbullet\enspace All of Mathlib and the benchmark's original imports are already loaded, so additional \texttt{import} lines are normally unnecessary. If you add or change imports, every check uses that exact current import header. An unknown Mathlib identifier is therefore usually a wrong or renamed lemma name: fix the name, and fully-qualify it (e.g. \texttt{Real.log}) if it is ambiguous. You MAY add \texttt{open} statements.
\par
\par\smallskip\noindent\textbf{Statement integrity}\par\nopagebreak
\par
\par\noindent\textbullet\enspace NEVER modify any given signature, name, type, or definition in the file.
\par\noindent\textbullet\enspace Only fill in the proof bodies (the \texttt{sorry}s); you may add your own helper lemmas/definitions.
\par\noindent\textbullet\enspace A file may contain SEVERAL target theorems / multiple \texttt{sorry}s --- prove EVERY one; the file is accepted only when ALL \texttt{sorry}s are filled.
\par
\par\smallskip\noindent\textbf{Avoiding verifier gaming}\par\nopagebreak
\par
\par\noindent\textbullet\enspace There is NO pre-made proof of the target sitting in the environment waiting to be found. Do not invent and \texttt{exact} a hopeful name like \texttt{<problem>\_true}, \texttt{<Author><Year>}, or \texttt{<problem>\_proved} --- a complete proof of the whole problem is not hiding under a guessed identifier. Find REAL supporting lemmas and prove the target yourself.
\par\noindent\textbullet\enspace \textbf{Do not metaprogram to gather information --- it cannot help.} You are in a sealed, controlled environment: the full Mathlib and every library the problem imports are ALREADY loaded. There is NO external source of information that would help you solve this, and everything required to prove the theorem is already in front of you and reachable in the provided environment. Attempting to read files or list directories (\texttt{IO.FS}, \texttt{System.FilePath}), spawn processes (\texttt{IO.Process}), or enumerate/inspect the environment by hand (\texttt{getEnv}, \texttt{env.constants}, custom \texttt{elab}/\texttt{syntax}/\texttt{run\_cmd}) returns nothing useful and only wastes the iteration. Spend your calls proving the theorem, not probing the environment. (Running genuine computations to test the statement numerically --- e.g. \texttt{\#eval} / \texttt{Id.run do} over small cases --- is fine and part of proving; the point is only that using code to inspect the environment or anything outside the proof will not help.)
\par
\par\smallskip\noindent\textbf{Disproofs}\par\nopagebreak
\par
\textbf{Prove OR disprove --- but a disproof requires real evidence, never convenience}: a target statement is not guaranteed to be true. Pivot to a disproof ONLY when you have a concrete counterexample you can construct and check in Lean, or a rigorous proof of the negation. NEVER pivot because the forward proof feels hard, long, or "impossible", because the problem is labelled open, or on the strength of an unverified recollection from the literature --- "I cannot prove it, so it must be false" is a fallacy that produces garbage. When (and only when) you have that evidence, rename the theorem to \texttt{<name>.disproof} and use one of these verifier-supported forms:
\par
\par\noindent\textbullet\enspace \textbf{Outer negation (preferred):} state the literal negation of the ENTIRE original theorem type. For \texttt{theorem foo (x\ldots{}) (h\ldots{}) : C}, write \texttt{theorem foo.disproof : \ensuremath{\neg} (\ensuremath{\forall} x\ldots{}, h\ldots{} \ensuremath{\to} C) := by ...}. Move every original parameter, typeclass binder, and hypothesis inside the negation. Preserve universe polymorphism; a concrete counterexample may need \texttt{ULift} witnesses to work at the target's universe levels.
\par\noindent\textbullet\enspace \textbf{Fully pushed negation (optional compatibility form):} when a direct witness is simpler, write the verifier's negation-normal form, such as \texttt{theorem foo.disproof : \ensuremath{\exists} x\ldots{} h\ldots{}, \ensuremath{\neg} C := \ensuremath{\langle}witnesses\ldots{}, by ...\ensuremath{\rangle}}. Continue pushing through leading \texttt{\ensuremath{\forall}}/\texttt{\ensuremath{\exists}}/\texttt{\ensuremath{\land}}/\texttt{\ensuremath{\lor}} connectives until the first atomic proposition.
\par
Both forms are kernel-checked and accepted. Prefer outer negation because it avoids reconstructing a pushed normal form and works reliably with dependent binders. If a pushed disproof gets a type mismatch, switch to the literal outer form.
\par
\par\smallskip\noindent\textbf{Soundness and acceptance}\par\nopagebreak
\par
The final proof is replayed and kernel-checked by SafeVerify.
\par\noindent\textbullet\enspace \texttt{sorryAx}, new \texttt{axiom} declarations, \texttt{unsafe}, and meta-programming (\texttt{elab}, \texttt{macro}, \texttt{syntax}, \texttt{run\_cmd}, \texttt{run\_tac}) cannot bypass proof obligations or final validation: SafeVerify replays the final theorem and rejects forbidden axiom dependencies. To DEFER a goal use plain \texttt{sorry} (it returns the goal state); \texttt{sorryAx} and new axioms add nothing toward an accepted proof.
\par\noindent\textbullet\enspace \texttt{decide} is kernel-checked and fine for exploration or the final proof. \texttt{native\_decide} may provide computational evidence during exploration, but it trusts compiled evaluation rather than the kernel, so SafeVerify rejects a final proof that depends on it.
\par\noindent\textbullet\enspace \texttt{sorry}/\texttt{admit} are fine for EXPLORATION --- they return the unsolved goal states --- but a FINAL proof containing them is never accepted (per requirement 2).
\par\noindent\textbullet\enspace A genuine proof may depend ONLY on the standard axioms \texttt{\{propext, Classical.choice, Quot.sound\}}; any other axiom dependency is rejected. (\texttt{decide} is allowed --- it is kernel-checked --- but prefer a real proof.)
\par\noindent\textbullet\enspace The theorem statement is re-checked every iteration: changing the signature/type is rejected. Only edit the proof body.
\par
\par\smallskip\noindent\textbf{Persistence}\par\nopagebreak
\par
Approach every problem with the curiosity, ambition, and stubborn persistence of a top researcher. \textbf{The difficulty of a problem --- whether it is "known", "open", "hard", or "long" --- is irrelevant to your job and must not enter your reasoning.} You have effectively unlimited iterations and compute, far more than any human has ever brought to bear on these problems. Some targets may be famous open conjectures; it does not matter. Like George Dantzig, who solved two famous unsolved problems in statistics because he arrived late to class and mistook them for homework, you may well solve a problem you would have thought impossible simply by not treating it as impossible. Treat every target as a solvable problem you have not cracked YET, and get to work.
\par
There is \textbf{no cap on iterations and compute is effectively unlimited}: the per-iteration turn budget is only a checkpoint, never a deadline, never a reason to rush, to submit a \texttt{sorry}, or to stop trying. Do not spend turns on no-op edits to "run out the clock", and do not submit an unchanged file --- every iteration must make genuine progress on the mathematics.
\par
\textbf{CRITICAL --- NEVER GIVE UP, AND NEVER CHEAT.} You do NOT need a complete proof in a single iteration, but you DO need to reach a full, \texttt{sorry}-free, honestly-verified proof of the real statement, however hard or long that takes. Keep grinding across iterations: when something is complex, break it into lemmas, sketch the structure with \texttt{sorry} to read the subgoals, prove the pieces, and iterate. When an honest proof looks long, that is a reason to SEARCH Mathlib for existing results that shorten it --- NOT a reason to give up, and NOT a signal that a shortcut around the mathematics must exist.
\par
\par\smallskip\noindent\textbf{Examples}\par\nopagebreak
\par
Target theorem: \texttt{theorem add\_zero (n : Nat) : n + 0 = n := sorry}
<example>
\textbf{First attempt - exploring tactics}
\par
Edit the file to:
\par\begingroup\ttfamily\obeylines
theorem add\_zero (n : Nat) : n + 0 = n := by
\hspace*{1em}simp
\endgroup\par
\par
(Next iteration would receive feedback about whether \texttt{simp} solved it or what goals remain)
</example>
\par
<example>
\textbf{Second iteration - incomplete proof with unsolved goals}
The code from the previous iteration builds successfully but goals remain: \texttt{\ensuremath{\vdash} n + 0 = n}. Therefore, the \texttt{simp} tactic alone doesn't solve this goal. A possible strategy now is to use induction to break this into base and inductive cases, leaving sorries to see what specific goals need to be proven in each case.
\par
Edit the file to:
\par\begingroup\ttfamily\obeylines
theorem add\_zero (n : Nat) : n + 0 = n := by
\hspace*{1em}induction n with
\hspace*{1em}| zero => sorry
\hspace*{1em}| succ n ih => sorry
\endgroup\par
\par
(This intentionally leaves sorries to get feedback about the base and inductive case goals)
</example>
\par
<example>
\textbf{Third iteration - Closing the goals}
Sorry \#1 at line 3, column 13 is \ensuremath{\vdash} 0 + 0 = 0, and sorry \#2 at line 4, column 13 is \ensuremath{\vdash} n + 1 + 0 = n + 1. Without these sorries, we have inspected the goal state in multiple points of the proof. Now we can proceed to close these goals.
\par
Edit the file to:
\par\begingroup\ttfamily\obeylines
theorem add\_zero\_1 (n : Nat) : n + 0 = n := by
\hspace*{1em}induction n with
\hspace*{1em}| zero => rfl
\hspace*{1em}| succ n ih => rw [Nat.add\_succ, ih]
\endgroup\par
</example>
\par
<example>
\textbf{Fourth iteration - fixing specific error}
The previous attempt's build failed with error at line 4: 'unknown identifier Nat.add\_succ'. The proof attempted to use a lemma that doesn't exist or isn't imported (the correct name is \texttt{Nat.succ\_add}). A reasonable next step is to search for the correct lemma name, although, after looking at it carefully, this is a case that \texttt{omega} (specialized for integer and natural linear arithmetic problems) can solve.
\par
Edit the file to:
\par\begingroup\ttfamily\obeylines
theorem add\_zero (n : Nat) : n + 0 = n := by
\hspace*{1em}omega
\endgroup\par
</example>

\end{promptquote}

\begin{promptquote}{Reflection LLM: detailed system prompt}
\par\smallskip\noindent\textbf{Role and inputs}\par\nopagebreak
\par
An LLM agent is trying to prove a theorem in Lean 4 and has not yet completed the full proof.
This prover agent works iteratively across many attempts, but it does NOT have access to its own history of previous attempts. Instead, it is your task to give it meaningful context condensing the experience gained from its previous attempts, so it can make progress on the next one.
When the prover takes its next attempt, it will be given both the current state of the proof (code and feedback) and the context that you prepare.
\par
You will receive a distilled summary of the prover agent's most recent attempt (its reasoning, its final code, and the verifier's feedback), followed by its interaction transcript: the system prompt, input, tool calls and results, and the prover's reasoning. The first reflection request contains the full transcript. If that request exceeds the context limit, redundant proof-file bodies may be replaced by explicit omission notes because the authoritative final code is already in the distilled summary. If the request still exceeds the context limit, interleaved prover turns and their associated tool results may also be omitted explicitly while inputs and the final turn remain.
The message also contains the previous context that you had prepared for the prover (its synthesized past experience). If the previous context was empty, this was the first attempt.
\par
\par\smallskip\noindent\textbf{Evidence discipline}\par\nopagebreak
\par
Write as a third-party account of the prover --- never as an oracle. The prover has no memory of its own attempts, so it reads whatever you write as authoritative. Therefore you MUST attribute every belief, conclusion, mathematical claim, and empirical observation to the prover itself --- "You believed\ldots{}", "You concluded\ldots{}", "You attempted\ldots{}", "You observed\ldots{}", "Your \texttt{\#eval}/\texttt{decide} showed\ldots{}", "You recalled from the literature\ldots{}". Do NOT restate the prover's beliefs as facts about the world. In particular, claims like "the theorem is true", "the theorem is false", "the answer is 0", or "such a set exists" are the prover's own conclusions and must be phrased as such ("You concluded the theorem was false, though the verifier has not confirmed this"), never as settled truth --- otherwise a wrong belief calcifies into an oracle verdict and the prover never revisits it. The ONLY two things you may state as plain fact are the things actually in the record: the exact contents of the proof file, and the verbatim feedback returned by the verifier. Concrete, verifier-observed tactical facts ("\texttt{omega} closed the goal"; "the goal at the remaining \texttt{sorry} is \texttt{\ensuremath{\vdash} \ldots{}}") may of course be stated directly, and you should keep such tactical guidance crisp and assertive.
\par
\par\smallskip\noindent\textbf{Strategy discipline}\par\nopagebreak
\par
For POSITIVE strategy --- new tactics, lemmas, or constructions to try next --- stick to the facts and refrain from proposing them yourself; the prover will draw its own conclusions from the record you provide. There is ONE mandatory exception, described next.
\par
\par\smallskip\noindent\textbf{Do not carry forward unsound or defeatist reasoning}\par\nopagebreak
\par
Every theorem here is meant to be proved with a sound, \texttt{sorry}-free proof, and the prover has effectively unlimited iterations to do it. If the record shows the prover reasoning toward any of the following, you must NOT restate that reasoning as a fact or a viable plan. Instead, record plainly --- in the prover's voice --- that it went down that path and that the path does not work, so it is not repeated:
\par
\par\noindent\textbullet\enspace \textbf{Cheating / gaming the verifier}: presenting \texttt{sorry}, \texttt{sorryAx}, \texttt{native\_decide}, or a new \texttt{axiom} as a finished proof; or guessing that a finished proof already exists under some identifier. State that it did not produce a valid final proof. If the verifier rejected that path, say so and quote the actual rejection feedback. Never present it as the way forward.
\par\noindent\textbullet\enspace \textbf{Giving up / defeatism}: submitting \texttt{sorry}, burning the turn budget on purpose, or calling the problem "open", "impossible", "too hard", "too long", or "infeasible in one session". NEVER assert, endorse, or repeat any such framing --- not even when the prover concluded it, and not even to "acknowledge reality"; the prover treats it as ground truth and stops. Record only the specific, concrete obstacle actually hit (e.g. "the inequality in lemma X remained unproved"; "no Mathlib lemma matching pattern Y was found").
\par\noindent\textbullet\enspace \textbf{False disproof}: pivoting to a \texttt{.disproof} from a hunch, a literature recollection, or the mere difficulty of the forward proof, rather than a verifier-confirmed counterexample or negation.
\par
Do not scold, threaten, or lecture the prover; simply decline to carry the unsound reasoning forward, and note neutrally that it did not work. Difficulty, length, and whether a problem is "open" are irrelevant and must never appear in the context you write.
\par
\par\smallskip\noindent\textbf{Interpret checkpoint submissions correctly}\par\nopagebreak
\par
An iteration always ends with a \texttt{submit}, but a submit does NOT mean the prover judged the proof complete. Watch for these cases (often the file still contains \texttt{sorry}), and do NOT read any of them as a deliberate, careless, or premature submit, nor write a lesson blaming the prover for "giving up" or "submitting without progress" --- treat the file as a work-in-progress checkpoint the prover continues next iteration:
\par
\par\noindent\textbullet\enspace "[NOTE: this submission was FORCED ...]" prepended to the feedback --- read the reason stated in the forced-submission note.
\par\noindent\textbullet\enspace "[NOTE: the proposer LLM call timed out ...]" prepended to the feedback --- the model's thinking hung and, after retries, the current file was auto-submitted with reasoning "thinking timeout"; an infrastructure timeout, not a decision by the prover.
\par\noindent\textbullet\enspace No NOTE, but the transcript shows the prover was told "You ended your turn without calling any tool, so this iteration is being closed" --- it stopped emitting tool calls with turns still to spare and was then asked to submit. This is the prover pausing, not finishing; do not treat its submit reasoning as a considered final verdict.
\par
\par\smallskip\noindent\textbf{Preserve useful progress}\par\nopagebreak
\par
Reflect on the current attempt in light of the previous context, and extract the key lessons that will help the prover avoid repeating this attempt's mistakes and all previous ones.
Preserve concrete, verifier-confirmed progress explicitly --- name the lemmas and definitions the prover has already proved and that compiled, and the exact Mathlib lemma names a search actually turned up --- so the prover does not re-derive or re-search them (it routinely loses proved helper lemmas and repeats dead-end name guesses across iterations).
Ensure the important information captured in the previous context is never lost, so the prover does not repeat old mistakes even many iterations later.
Keep the context concise and to the point, with the goal of maximizing the prover's performance.
\par
\par\smallskip\noindent\textbf{Output}\par\nopagebreak
\par
Your output will be given to the prover agent in the next iteration. The harness will embed it as the content inside a \texttt{<problem-experience>} block. Return ONLY the body of that block. Do NOT output \texttt{<problem-experience>} tags, the surrounding message template, or the introductory sentence; the harness adds them exactly once.

\end{promptquote}

\subsection{Lean feedback and error handling}
\label{app:lean-recovery}

Lean errors and process failures receive different treatment (Table~\ref{tab:lean-recovery}).
A successful Lean restart lets the same proof attempt continue; it does not start a new independent model attempt. After a connection failure, \hone{} repeats the same Lean command without resampling the model response. Interrupted executions can resume from saved proof state within the same benchmark attempt; they are not additional samples for pass@$k$.

\begin{table}[H]
\centering
\caption{\textbf{How \hone{} handles Lean errors and process failures.} Proof errors
are returned to the model for revision. Process failures trigger a restart or end
the current execution, depending on where they occur. Default timeouts are 600 seconds
for Lean snippets and 1,200 seconds for the working file.}
\label{tab:lean-recovery}
\footnotesize
\setlength{\tabcolsep}{4pt}
\renewcommand{\arraystretch}{1.15}
\begin{tabularx}{\textwidth}{@{}p{0.17\textwidth}p{0.25\textwidth}X p{0.18\textwidth}@{}}
\toprule
What goes wrong & What \hone{} does & What the model sees & How the attempt continues \\
\midrule
Lean error or open goal
  & Keep the edit and check the file.
  & Lean's errors and remaining goals.
  & The model can revise the proof in the same iteration. \\
REPL connection loss
  & Restart Lean and repeat the same command once.
  & The command's result if it succeeds; otherwise a failure message from the tool.
  & If the command succeeds, the same model interaction continues. \\
REPL timeout
  & Restart Lean without repeating the command that timed out.
  & A timeout message from the tool.
  & Edit checks restore the previous file; scratch execution allows one failure. A timeout before submission interrupts the execution. \\
Process failure during an edit check
  & Restore the file to its state before the edit and check it after restarting Lean.
  & A message that the edit was rejected and already undone.
  & Continue if the restored file checks; otherwise interrupt the execution. \\
Process failure during scratch-code execution
  & Leave the working file unchanged. Allow one failure; stop after two in a row.
  & A warning that the scratch computation ran out of time or memory.
  & After the first failure, the model can try again. A second consecutive failure interrupts the execution. \\
Initial health check fails
  & Stop if Lean cannot check the original file before proving begins.
  & Nothing: no model call has been made.
  & Proving can begin once the Lean service is working. \\
Errors, open goals, or process failure before submission
  & Check the proof before SafeVerify. Reject proofs with errors or open goals; stop the execution if Lean fails.
  & Lean's errors or remaining goals. If the process fails, the execution ends before feedback reaches the model.
  & The model can revise a rejected proof. A process failure interrupts the execution, which can resume from saved state. \\
SafeVerify timeout
  & Stop compilation in the validation sandbox and reject the proof.
  & A message that validation timed out.
  & Give the model another iteration to revise the proof. \\
\bottomrule
\end{tabularx}
\end{table}

A deterministic Lean failure that recurs on three consecutive executions is counted
as an unsolved problem. Interruptions caused by the model provider, compute cluster,
or tool service can instead be resumed from the saved proof state. Unless otherwise
stated, interrupted and unfinished problems remain in the evaluation denominator.

\clearpage

\section{Proof Validation, Evaluation, and Cost Accounting}
\label{app:evaluation}

This appendix defines which proofs count as solutions, how inference costs and
budget-limited solve rates are computed, and how uncertainty is estimated. It also
specifies differences in cost reporting that affect comparisons with published systems.

\subsection{Benchmark revisions and Lean environments}
\label{app:benchmark-versions}

Table~\ref{tab:benchmark-versions} summarizes the source revisions and Lean/Mathlib
environments in the setup recipes for the benchmarks used in this paper.
Putnam100 and the 2025 slice share the PutnamBench source; the separate Numina
formalization is listed independently.

\begin{table}[H]
\centering
\caption{\textbf{Benchmark sources and Lean environments.}
Source revisions and Mathlib releases follow the setup recipes;
Lean versions are cross-checked against run metadata. Dates are source commit dates.}
\label{tab:benchmark-versions}
\footnotesize
\setlength{\tabcolsep}{4pt}
\renewcommand{\arraystretch}{1.15}
\begin{tabularx}{\linewidth}{@{}>{\raggedright\arraybackslash}p{0.31\linewidth}>{\raggedright\arraybackslash}Xll@{}}
\toprule
Benchmark & Source revision & Lean & Mathlib \\
\midrule
PutnamBench (including Putnam100 and the 2025 slice)
& \href{https://github.com/trishullab/PutnamBench/commit/3c9b9268ebd132d482938b8e9d242dd71fc1ebc0}{\texttt{3c9b9268}} (2026-07-23)
& 4.27.0 & 4.27.0 \\
FATE-X
& \href{https://github.com/frenzymath/FATE-X/commit/a6fb49ee85fd05bb8ce9fc66e1c66797b5a0592d}{\texttt{a6fb49ee}} (2026-02-23)
& 4.28.0 & 4.28.0 \\
FormalQualBench
& \href{https://github.com/math-inc/FormalQualBench/commit/efaa113c6a00a79e92842ce541b407d7695d7699}{\texttt{efaa113c}} (2026-03-19)$^*$
& 4.28.0 & 4.28.0 \\
FC100-solved
& \href{https://github.com/google-deepmind/formal-conjectures/commit/7862bf6a6976c26acb086b6b2cec12ad1d92ce5d}{\texttt{7862bf6a}} (2026-04-29)
& 4.27.0 & 4.27.0 \\
Erd\H{o}s problems (APN comparison)
& \texttt{7862bf6a} (2026-04-29)
& 4.27.0 & 4.27.0 \\
Putnam 2025 (Numina)
& \href{https://github.com/project-numina/Numina-Putnam2025/commit/60d33c8ba19af905bd731e938ebde1c5b8c76519}{\texttt{60d33c8b}} (2026-01-20)
& 4.26.0 & 4.26.0 \\
IMO 2026
& \href{https://github.com/AxiomMath/IMO2026/commit/c5a6a089d06d3619afe7ff45c5ccab9e2a30d5d2}{\texttt{c5a6a089}} (2026-07-17)$^*$
& 4.29.1 & 4.29.1 \\
\bottomrule
\end{tabularx}
\par\smallskip
\begin{minipage}{\linewidth}
\scriptsize
The final two-problem PutnamBench follow-up uses
\href{https://github.com/trishullab/PutnamBench/commit/b9e18f601faeb68e52726a41a275ca65ada978af}{\texttt{b9e18f60}}
(2026-08-26), with the same Lean/Mathlib versions
(Appendix~\ref{app:putnam-followup}). The FATE-X revision is pinned by the comparison setup.
$^*$These setup scripts fetch upstream without a commit pin; the listed revisions
were resolved on September~25, 2026.
\end{minipage}
\end{table}

\paragraph{Formal Conjectures snapshot and open status.}
The FC100-solved and Erd\H{o}s/APN setups use Formal Conjectures revision
\href{https://github.com/google-deepmind/formal-conjectures/commit/7862bf6a6976c26acb086b6b2cec12ad1d92ce5d}{\texttt{7862bf6a}},
dated April~29, 2026, with Lean/Mathlib~4.27.0. This source date is distinct from
the May~6 benchmark release.
We use the original solved subset, \texttt{FC100Solved}, subsequently renamed \texttt{FC100SolvedSet1} upstream.
The separate evaluation of 236 open targets is
described in Appendix~\ref{app:open-solved}.

\subsection{Formal validation}
\label{app:formal-validation}

A compiling file need not prove the intended theorem: it may contain \texttt{sorry},
change assumptions, or introduce axioms. We use SafeVerify~\citep{safeverify} to compare
compiled submissions with an immutable challenge, check target declarations and permitted
axioms, and replay new declarations through Lean's kernel. Imported modules remain part
of the trusted environment.

Comparator~\citep{comparator} also checks statement agreement, permitted axioms,
and kernel acceptance, and integrates sandboxed compilation of the challenge
and submission. AXLE's \texttt{verify\_proof} tool~\citep{axle2026verifyproof} offers similar checks, including optional verification of a statement's negation, through a hosted API. SafeVerify instead takes already-compiled \texttt{.olean} files,
so the harness must isolate compilation and protect the challenge and imported
dependencies~\citep{safeverify}. We chose SafeVerify for its disproof support,
which allows a fixed conjecture to be settled by proving it or its negation,
and because Comparator's Landlock/landrun sandbox was not supported in our
execution environment.
We built SafeVerify against each
benchmark's toolchain and implemented the isolated compilation procedure in
Table~\ref{tab:validation-boundary}.

\begin{table}[H]
\centering
\caption{\textbf{Independent validation of submitted source.} The same boundary applies to API provers and coding agents. Only the designated source file crosses from the agent workspace into validation; the agent's compiled artifacts are never trusted.}
\label{tab:validation-boundary}
\footnotesize
\setlength{\tabcolsep}{4pt}
\renewcommand{\arraystretch}{1.12}
\begin{tabularx}{\textwidth}{@{}p{0.22\textwidth}X@{}}
\toprule
Stage & Check \\
\midrule
Trusted challenge & Compile the original statement with fixed Lean and library versions in an isolated environment; retain its compiled artifact in storage inaccessible to the prover. \\
Submission compile & Compile the submitted source in a separate, fresh, networkless sandbox against read-only dependencies, preserving the module's relative path. \\
Kernel validation & Compare the original and submitted compiled artifacts using SafeVerify in a trusted Lean project. Accept only when the verifier reports success and exits without error. \\
Proof or disproof & Accept a proof of the target theorem. Where disproofs are enabled, also accept a \texttt{.disproof} declaration proving the supported negation of that theorem. \\
\bottomrule
\end{tabularx}
\end{table}

\paragraph{Comparison cutoff.}
\label{app:comparison-cutoff}
We stopped adding results from other methods to our comparison set after August~27, 2026.
All state-of-the-art claims in this paper are relative to results publicly available by that date; later work is discussed only as related or concurrent work.
For PutnamBench, the \href{https://github.com/trishullab/PutnamBench/blob/3749f97a24d99dd5c771eb3bc435b8695b16475b/docs/results.json}{leaderboard snapshot of August~27, 2026} includes Aleph's and Humanfia's 672-solve results and Goedel-Architect's July update.
Source access dates after the cutoff reflect checks made while writing, rather than an extension of the comparison period.

\subsection{Inference budgets and cost curves}
\label{app:budget-comparisons}
An attempt can denote a single model call, an iterative repair session, or a complete
multi-stage pipeline. For example, Goedel-Architect allows up to 16 blueprint refinements
within one PutnamBench attempt~\citep{goedelarchitect2026}. Matching attempt counts
therefore need not match inference cost. Total token counts are also insufficient:
uncached input, cached input, and output have different costs, with output tokens
costing up to $100\times$ as much as cached input in our price cards.
Ideally, we would compare inference
FLOPs directly, but closed-model providers rarely disclose the information needed
to measure them. We therefore use inference cost as a practical proxy and compare
cost with the model held fixed to assess
harness differences; input, cached input, cache writes, and output are priced separately
using the reference rates in Table~\ref{tab:model-token-prices}, including applicable
long-context rates. These API-equivalent costs quantify token usage under a
specified price card, not actual expenditure.

\begin{table}[H]
\centering\footnotesize
\setlength{\tabcolsep}{4pt}
\caption{\textbf{Reference token prices, in US dollars per million tokens.}
Public API prices are reported by \href{https://artificialanalysis.ai/}{Artificial Analysis}
(accessed September 24, 2026); GPT-5.6~Sol retains its
\href{https://artificialanalysis.ai/articles/gpt-5-6-has-landed}{July 9 launch rates}.
Leanstral uses Mistral Small~4 as a pricing proxy; DeepSeek uses the fixed
April 26, 2026 reference card described in Appendix~\ref{app:deepseek-costs}.
Input denotes uncached input; cache reads and writes are counted separately.
The cache-write columns distinguish five-minute and one-hour retention;
a dash denotes no separate charge in our accounting.
Gemini~3.1~Pro's input-length tier applies to the entire call. Reasoning tokens count as output.
Reported curves and costs retain the experimental price cards.
Costs quoted from other studies retain their reported accounting.}
\label{tab:model-token-prices}
\begin{tabularx}{\textwidth}{@{}Xrrrrr@{}}
\toprule
Model and configuration & Input & Cache read & \multicolumn{2}{c}{Cache write} & Output \\
\cmidrule(lr){4-5}
 & & & 5 min & 1 hour & \\
\midrule
Claude Haiku~4.5 & 1.00 & 0.10 & 1.25 & 2.00 & 5.00 \\
Claude Opus~4.5, 4.6, 4.7, 4.8 & 5.00 & 0.50 & 6.25 & 10.00 & 25.00 \\
Gemini~3~Flash & 0.50 & 0.05 & -- & -- & 3.00 \\
Gemini~3.1~Pro ($\leq$200k input tokens) & 2.00 & 0.20 & -- & -- & 12.00 \\
Gemini~3.1~Pro ($>$200k input tokens) & 4.00 & 0.40 & -- & -- & 18.00 \\
GPT-5.2 & 1.75 & 0.175 & -- & -- & 14.00 \\
GPT-5.4 & 2.50 & 0.25 & -- & -- & 15.00 \\
GPT-5.5 and GPT-5.6~Sol & 5.00 & 0.50 & -- & -- & 30.00 \\
Leanstral~1.5 (Mistral Small~4 proxy) & 0.15 & 0.015 & -- & -- & 0.60 \\
DeepSeek~V4~Flash (API-equivalent estimate) & 0.14 & 0.0028 & -- & -- & 0.28 \\
\bottomrule
\end{tabularx}
\end{table}
For problem $i$, let $c_i$ be its total API-equivalent cost and $a_i$ its cost at first
acceptance, with $a_i=\infty$ if unsolved. At per-problem cap \$$B$, the solve count is
$\sum_i\mathbf{1}[a_i\le B]$. Figures with a budget axis plot this count against
\$$B$; figures with an average-cost axis plot it against
$\frac{1}{n}\sum_i\min(c_i,B)$. Where first-acceptance cost is unavailable,
the final cost of a successful attempt is used as its proxy. Cost bars average
all recorded problems, including failures. A nominal cap is an allocation limit;
recorded final cost can exceed it when the last model call crosses the limit.
Every curve ends at a finite budget. We stopped runs when further spending became
unreasonable under our time and compute limits, or when Lean failed in ways the harness
could not recover from, such as repeated REPL timeouts; the latter call for a better
Lean interface rather than a different loop.

\paragraph{Matching prior work's reporting conventions.}
To compare with published results, we follow the corresponding benchmark's
reporting conventions where needed. In Table~\ref{tab:complex}, costs are campaign totals for Putnam 2025, per-problem means
for PutnamBench and IMO, and means over successful problems for FormalQualBench.
Numina reports a budget cap rather than a cost derived from token usage. Putnam and IMO times are summed
across problems; FormalQualBench times average successes, and Humanfia's IMO time includes
only API time. Putnam 2025 uses Numina's statements, whereas the PutnamBench 2025 slice
uses separate formalizations. Our FormalQualBench Opus~4.6 result is the best of four
seeds (7, 6, 8, and 6 solves); the leaderboard model attribution to Opus~4.6 is inferred.
The best@6 row on PutnamBench's 2025 slice gives the order of magnitude a favourable
seed can reach among six attempts; its cost counts successful attempts only. We report it
next to pass@1 and pass@4, and the four FormalQualBench seeds individually, so that the
spread is visible; we draw no conclusion from the best seed alone.
The cost measure therefore changes across tables and figures: to compare with a published
number, we adopt that work's convention, such as campaign totals, budget caps, or cost
conditional on success. Wherever our data allow, we also report the full cost including
failed attempts, and we claim comparable or lower cost at matched models rather than
uniform cost dominance across systems.

\subsection{PutnamBench results and published cost conventions}
\label{app:putnam-followup}
Table~\ref{tab:putnam672-plot-points} reports solve percentages so that benchmarks with
different denominators can be distinguished. Published state-of-the-art results often
combine several campaigns that re-attempt residual problems, such as Aleph's budget
re-launches~\citep{alephprover2026reports}. Where possible, we report a single campaign
without re-running residual problems; the one exception is the targeted completion of the
last two problems described below. Published costs are not uniformly measured:
Goedel-Architect reports campaign totals (\$294 and \$985.67, divided by 672 here),
whereas Aleph's first three points report cost per successful problem~\citep{alephprover2026reports}. Aleph's final
result reports \$73.98 per problem, excluding attempts preceding formalization fixes;
the figure rounds this to \$74. Humanfia reports cost per benchmark problem. Its current
results attribute 672 solves at \$44.50 to GPT-5.6~Sol; the earlier GPT-5.5 result is
670 at \$44.60 (repository accessed 14 September 2026).
Hilbert's approximate \$244 cost is an estimate attributed to
\citet[Table~2]{goedelarchitect2026}, based on pass@1 statistics. It is not measured
cost for the 70\% headline result: Hilbert did not track compute for that multi-attempt
run, and its separate compute-tracked pass@1 run achieved 55.9\%~\citep[Appendix~A.6]{hilbert2025}.

\begin{table}[H]
\centering\footnotesize
\setlength{\tabcolsep}{3pt}
\caption{\textbf{PutnamBench results underlying Figures~\ref{fig:teaser} (left) and~\ref{fig:deepseek-reflection}.}
Solve rates use 672 problems except Hilbert (462/660, 70.0\%), DeepSeek~V4~Flash
with TIR (109/200, 54.5\%), and Aleph's first result (500/660, 75.8\%).
The figure places Hilbert and TIR at equivalent rates on its 672-problem axis;
Aleph's first point retains its published count of 500. Dollar values are per-problem
means unless qualified above; Hilbert's cost is an attributed estimate, and TIR
reports no cost. Our DeepSeek values are estimated API-equivalent costs.}
\label{tab:putnam672-plot-points}
\begin{tabular}{@{}>{\raggedright\arraybackslash}p{0.48\linewidth}lrr@{}}
\toprule
System / model & Setting & Solved (\%) & \$/problem \\
\midrule
Ours + GPT-5.5 & Capped & 87.5 & 5.28 \\
 & Capped & 95.5 & 6.63 \\
 & Endpoint & 99.7 & 9.17 \\
 & + 8 attempts & 100.0 & 17.86 \\
\addlinespace[2pt]
Ours + DeepSeek V4 Flash & Capped & 69.5 & 0.40 \\
 & Capped & 75.6 & 0.46 \\
 & Capped & 82.4 & 0.69 \\
 & Endpoint & 89.0 & 1.08 \\
\addlinespace[2pt]
Ours + DeepSeek V4 Flash, no reflection & Capped & 69.5 & 0.57 \\
 & Capped & 75.6 & 0.78 \\
 & Endpoint & 82.4 & 1.96 \\
\addlinespace[2pt]
Ours + Gemini 3.1 Pro & Capped & 95.5 & 6.32 \\
 & Capped & 97.6 & 7.14 \\
 & Capped & 99.7 & 11.44 \\
\addlinespace[2pt]
Goedel-Architect + DeepSeek V4 Flash & pass@1 & 75.6 & 0.44 \\
\addlinespace[2pt]
Goedel-Architect + DeepSeek V4 Flash/Pro & Proof-seeded @4 & 88.8 & 1.47 \\
\addlinespace[2pt]
Goedel-Architect + Gemini 3.1 Pro Preview / DeepSeek V4 Pro & July 2 & 95.5 & 5.80 \\
\addlinespace[2pt]
Humanfia + GPT-5.5 & Reported & 99.7 & 44.60 \\
\addlinespace[2pt]
Humanfia + GPT-5.6 Sol & Reported & 100.0 & 44.50 \\
\addlinespace[2pt]
Aleph Prover & Reported & 75.8 & 23.00 \\
 & Reported & 94.8 & 54.00 \\
 & Reported & 99.4 & 68.00 \\
 & Reported & 100.0 & 74.00 \\
\addlinespace[2pt]
Hilbert: Gemini 2.5 Pro + Goedel-Prover-V2-32B & Published result & 70.0 & $\approx244.00$ \\
\addlinespace[2pt]
DeepSeek V4 Flash + TIR (Goedel-Architect run) & pass@32 & 54.5 & -- \\
\bottomrule
\end{tabular}

\end{table}

\paragraph{Targeted completion of the last two problems.}
The GPT-5.5 continuation ran eight replicas each on \texttt{putnam\_2013\_a5} and
\texttt{putnam\_2021\_a6} using benchmark revision \texttt{b9e18f6} and Lean~4.27, obtaining seven
and one SafeVerify-accepted proofs, respectively. PutnamBench has corrected formalizations since
our original campaign, including that of \texttt{putnam\_2013\_a5}; these corrections
also allowed other systems to reach 672 solves~\citep{alephprover2026reports,humanfia2026}.
The 672 result therefore combines our original campaign with the corrected statements. For each problem $i$,
we estimate parallel stopping cost as $\sum_j\min(c_{ij},s_i)$, where $c_{ij}$ is
replica $j$'s final cost and $s_i$ the cheapest successful replica's cost. This adds
\$5{,}839.36, yielding \$17.86 per problem across 672 problems. It assumes cancellation
at equal dollar cost, not equal elapsed time. Including the full \$7{,}938.75 continuation
instead gives \$20.99 per problem.

\subsection{Estimated DeepSeek API costs}
\label{app:deepseek-costs}
DeepSeek~V4~Flash was served locally without recorded cache-hit counters.
We reconstruct cache reuse from unchanged prompt prefixes in the traces,
separately for each agent role and resetting at resume boundaries. This assumes
that reusable prefixes remain cached. We price the recorded input and output
tokens, including reflection and reasoning, at the fixed April~26 rates in
Table~\ref{tab:model-token-prices}. Goedel-Architect retains its reported costs;
its price card and cache usage are unspecified.
This gives 598/672 solves at an estimated \$1.08 per problem, with 92.27\% reconstructed
cache reuse.

\begin{table}[H]
\centering\footnotesize
\setlength{\tabcolsep}{4pt}
\caption{\textbf{Fraction of input tokens read from cache.}
DeepSeek values are reconstructed estimates; others use provider-reported counters.
Models use \hone{}
unless stated otherwise. A dash denotes no listed run; n.r.\ means no cache-read
counters were recorded.}
\label{tab:cache-rates}
\begin{tabular}{@{}lrrr@{}}
\toprule
Model / harness & PutnamBench & FC100-solved & FATE-X \\
\midrule
DeepSeek V4 Flash (estimated) & 92.26\% & 96.38\% & 96.15\% \\
GPT-5.5 & 79.09\% & -- & -- \\
GPT-5.6 Sol & -- & 93.77\% & 96.21\% \\
Gemini 3.1 Pro & 55.39\% & 61.21\% & 69.96\% \\
Claude Opus 4.5 & -- & -- & 91.64\% \\
Claude Opus 4.8 & -- & -- & 91.50\% \\
Claude Code + Opus 4.8 & -- & -- & 97.21\% \\
Leanstral 1.5 & -- & -- & n.r. \\
\midrule
DeepSeek, no reflection (estimated) & 95.49\% & -- & -- \\
\bottomrule
\end{tabular}

\end{table}

\subsection{Statistical significance and uncertainty}
Many comparisons in this paper rest on a single run per configuration. Repeating every
run would multiply an already large inference cost, and we judge single runs sufficient
for the claims they support, because most of them are existence claims: that a simple
loop \emph{can} reach a given score at a given cost. Since every accepted proof is formally
verified, one run establishes such a claim; for example, a single run shows that \hone{}
with GPT-5.5 can solve 670 of the 672 PutnamBench problems. Where a claim requires statistical significance, we repeat runs or resample problems and report 95\% confidence intervals, as in the
tool ablation (Figure~\ref{fig:tools-ablation}), the Erd\H{o}s prompt comparison
(Figure~\ref{fig:erdos-prompt-main}), and the reflection-input costs
(Table~\ref{tab:reflector-costs}).

\paragraph{Ablation effect sizes.}
Figure~\ref{fig:tools-ablation} summarizes both coverage and cost. Its tool
comparison rewards solving earlier: if $S(b)$ is the fraction solved by cost $b$,
we summarize the curve by $E(B)=\frac{100}{B}\int_0^B S(b)\,db$. Equivalently, each
problem solved at cost $a_i\le B$ earns $100(1-a_i/B)$ and an unsolved problem earns
zero; the score averages all 100 problems. The API-equivalent cost of a successful run is our proxy
for first-acceptance cost. We compare configurations by averaging their score differences
on the same problems. Upper boxes use the working-file-only configuration with the detailed
prompt as reference; lower boxes use all tools with the detailed prompt.

The 95\% percentile intervals use 50,000 bootstrap resamples of whole paired problems,
sharing resampling indices across configurations within a benchmark.

For the Erd\H{o}s prompt comparison in Figure~\ref{fig:erdos-prompt-main}, we compute
95\% bootstrap intervals by resampling the observed runs with replacement:
16 runs for the detailed-prompt reference and ten for each other configuration.
Each sampled run retains its outcomes on all six problems. Each resample
recomputes pass@$k$ at the \$150 per-attempt cap, with crashes and missing results
counted as unsolved.

\subsection{Proof screening}
We recover AxProverBase proofs from traces' final candidates. Removing
\texttt{native\_decide} removes 35 accepted proofs across nine runs. Screened counts
are used in the aggregate matched-cap comparison in Section~\ref{sec:comparison};
Figure~\ref{fig:teaser} (right) retains unscreened counts. SafeVerify accepts our proofs and rejects \texttt{sorryAx}, new axioms, and evaluation via \texttt{native\_decide}.

\clearpage

\section{Open-Problem Proofs and Statement Corrections}
\label{app:open-solved}

\subsection{Statements proved}
Table~\ref{tab:open-statements} states the eight mathematical results reported in
Section~\ref{sec:benchmarks} and identifies their Lean proof files. Written on the
Wall~II Conjectures~100 and~314 were proved after correcting their statements;
the other six concern the named distributed targets. Each result concerns the specific variant and formal statement listed in the table.

\begin{table}[H]
\centering
\caption{\textbf{Eight Formal Conjectures targets proved by \hone{}.} Each row states the result proved in the corresponding Lean file, listed beneath it. $^\dagger$The distributed statement was repaired before it was proved: Conjecture~100 took the degree norm of $G$ rather than of its complement, and Conjecture~314 used a different induced-path invariant.}
\label{tab:open-statements}
\small
\setlength{\tabcolsep}{4pt}
\renewcommand{\arraystretch}{1.15}
\begin{tabularx}{\textwidth}{@{}>{\raggedright\arraybackslash}p{0.20\linewidth}X@{}}
\toprule
Target & Theorem proved \\
\midrule
 
Erd\H{o}s~138, variant \texttt{dvd\_two\_pow}
& Let $W(k)$ be the two-colour van der Waerden number, the least $N$ such that
every two-colouring of $\{1,\dots,N\}$ contains a monochromatic $k$-term
arithmetic progression. Then $W(k)/2^{k}\to\infty$.
\newline{\scriptsize\texttt{Erdos138\_dvd\_two\_pow\_solution.lean}} \\
 
Erd\H{o}s~337, variant \texttt{ruzsa\_turjanyi}
& For every asymptotic additive basis $A\subseteq\mathbb{N}$ of density zero,
that is with $|A\cap[1,N]|=o(N)$, the ratio
$|(A+A)\cap[1,2N]|\,/\,|A\cap[1,N]|$ tends to infinity.
\newline{\scriptsize\texttt{Erdos337\_ruzsa\_turjanyi\_solution.lean}} \\
  
OEIS A108081
& Let $X$ be the smallest set of words over $\mathbb{Z}$ containing $[0]$ and
closed under $(u,v)\mapsto l(u)\,v$ and $(u,v)\mapsto u\,r(v)$, where $l$ and
$r$ reverse a word and shift its letters by $-1$ and $+1$. For every $n\ge1$,
the number of words of length $n$ in $X$ equals $a(n-1)$, where
$a(n)=\sum_{k=0}^{n}\binom{n+k-1}{k}F_{n-k+1}$.
\newline{\scriptsize\texttt{OeisA108081\_count\_words\_in\_x\_is\_a\_shifted.lean}} \\
 
OEIS A211417
& For $a(n)=(30n)!\,n!\big/\bigl((15n)!\,(10n)!\,(6n)!\bigr)$, every prime $p\ge5$ and
every $k\ge1$ satisfy $p^{3k}\mid a(p^{k})-a(p^{k-1})$ (Bala's
supercongruence).
\newline{\scriptsize\texttt{OeisA211417\_supercongruence\_solution.lean}} \\
 
OEIS A022030
& Let $a(0)=4$, $a(1)=16$ and, with truncated division,
$a(n+2)=\bigl(a(n+1)^{2}+a(n)-1\bigr)/a(n)-1$ for even $n$ and
$a(n+2)=a(n+1)^{2}/a(n)+1$ for odd $n$. Then
$a(n)=4a(n-1)-a(n-3)+a(n-4)$ for every $n\ge4$.
\newline{\scriptsize\texttt{OeisA22030\_conjecture.lean}} \\
 
Open Quantum Problem~35, \texttt{ame\_9\_10\_open}
& There exists a unit vector $\psi\in\left(\mathbb{C}^{10}\right)^{\otimes9}$
whose reduced density matrix on any four of the nine parties is the maximally
mixed state; that is, an AME$(9,10)$ state exists.
\newline{\scriptsize\texttt{OpenQuantumProblem35\_ame\_9\_10\_open\_solution.lean}} \\
 
Written on the Wall~II, Conjecture~100$^\dagger$
& For every connected graph $G$,
$\alpha(G)\le\bigl\lceil\bigl(\max_{v} l(v)+\tfrac12\|\deg G^{c}\|_{2}\bigr)/2\bigr\rceil$,
where $\alpha$ is the independence number, $l(v)$ is the independence number of
the neighbourhood of $v$, and the last term is the $\ell_{2}$ norm of the degree
sequence of the complement.
\newline{\scriptsize\texttt{WrittenOnTheWallII\_GraphConjecture100\_conjecture100.lean}} \\
 
Written on the Wall~II, Conjecture~314$^\dagger$
& Every connected triangle-free graph on at least two vertices whose largest
induced path has at most four vertices is well totally dominated: all of its
minimal total dominating sets have the same cardinality.
\newline{\scriptsize\texttt{WrittenOnTheWallII\_GraphConjecture314\_conjecture314.lean}} \\
 
\bottomrule
\end{tabularx}
\end{table}

\subsection{Misformalizations and editable answers}
\label{app:misformalizations}

Kernel checking establishes a proposition, but does not establish that it
expresses the intended problem. Table~\ref{tab:misformalizations} summarizes
statement defects identified in our Formal Conjectures audit and the benchmark
repairs. An editable answer creates a separate problem: when a submission can
change a definition referenced by the target, it can make the target reflexive
or vacuous. Such outputs do not count as solutions. We fix answers or quantify
witnesses satisfying the intended mathematical constraints, and exclude exact-answer targets when that transformation
would discard the mathematical question.

The smooth four-dimensional Poincar\'e variant illustrates a missing hypothesis. Its \texttt{SmoothConjectureFor} predicate omitted
\texttt{T2Space}, allowing non-Hausdorff smooth manifolds homotopy equivalent to
$\mathbb S^4$ but not homeomorphic to it. A concurrent upstream correction adds the
Hausdorff assumption and a Lean counterexample to the old statement~\citep{poincareHausdorffFix2026}.

\begin{table}[H]
\centering
\caption{\textbf{Statement defects and benchmark repairs.} Repairs here refer to
our benchmark statements, rather than the status of the upstream repository.
A repair does not imply that the repaired problem was solved: among these
rows, Written on the Wall~II Conjectures~100 and~314 contribute to
Table~\ref{tab:open-statements}. Editable-answer defects can coexist with a
mathematical misformalization.}
\label{tab:misformalizations}
\small
\setlength{\tabcolsep}{4pt}
\renewcommand{\arraystretch}{1.12}
\begin{tabularx}{\textwidth}{@{}>{\raggedright\arraybackslash}p{0.20\linewidth}XX@{}}
\toprule
Target & Defect & Correction or scoring decision \\
\midrule
Erd\H{o}s~507
& Signed triangle area and repeated vertices do not express the intended
minimum nonnegative area of three distinct points.
& Proofs of the original targets do not establish the intended geometric result. \\
Green~16
& One natural-number answer is used for every $N$, although the extremal
quantity depends on $N$.
& Define the extremal function of $N$ and state the asymptotic bound. \\
Erd\H{o}s~566/567
& The size Ramsey number counts edges of an arbitrary host, whereas the
source asks for the ordinary Ramsey number.
& Use the least number of vertices of a complete host. \\
Lam--Litt
& A chosen denominator can vanish at the initial point, making the
cleared-denominator identity degenerate.
& Require that denominator to be nonzero at the initial point. \\
Written on the Wall~II, 100/160/314
& The extraction loses a complement, substitutes a cycle count for a binary
indicator, or uses average distance for induced-path length.
& Restore the complement-degree norm, the $C_4$-free indicator, and the largest
induced-path order, respectively. \\
Open Quantum Problem~13
& Editable answers do not fix the proposed maximum number of mutually
unbiased bases.
& State the fixed dimension-six conjecture \texttt{IsMaxMUBCount 6 3}. \\
Other editable-answer targets
& Choosing an answer can make an equality reflexive or an asymptotic claim
vacuous; choosing and refuting an answer is also uninformative.
& Fix the answer or quantify a substantive witness; exclude tautological
exact-answer reformulations and disable disproof mode for answer-hole targets. \\
\bottomrule
\end{tabularx}
\end{table}

\clearpage

\clearpage

\section{Additional Results for Section 3}
\label{app:comparison-extra}
This appendix reports the per-problem results, model comparisons, and analyses
of failed attempts supporting Section~\ref{sec:results}.

\begin{table}[t]
\centering
\caption{\textbf{Full comparison including measured times.}
The results of Table~\ref{tab:complex}, with the Time column retained.
Cap denotes the time limit; Time reports measured time, summed across problems for Putnam and IMO and averaged over successes for FormalQualBench.}
\label{tab:complex-with-time}
\footnotesize
\setlength{\tabcolsep}{3.5pt}
\renewcommand{\arraystretch}{1.02}
\resizebox{\textwidth}{!}{%
\begin{tabular}{@{}llcccc@{}}
\toprule
Benchmark & System + model & Time cap & Solved & Cost & Time \\
\midrule
\multirow{4}{*}{\shortstack[l]{Putnam 2025\\(Numina formalization)}}
  & Numina, no informal prover, Opus~4.5 & -- & 4/12 & -- & -- \\
  & \textbf{\hone{} + Opus~4.5} & \textbf{--} & \textbf{7/12} & \$1{,}429 total & 6{,}122 min \\
\cmidrule(lr){2-2}
  & \textbf{Numina, full system, Opus~4.5 + Gemini~3~Pro + GPT-5.2} & \textbf{--} & \textbf{12/12} & \$1{,}800 cap & 3{,}889 min \\
  & \textbf{\hone{} + GPT-5.2} & \textbf{--} & \textbf{12/12} & \$565 total & 9{,}636 min \\
\midrule
\multirow{6}{*}{\shortstack[l]{PutnamBench's\\2025 slice }}
  & \hone{} + Opus~4.7 & -- & 11/12 & \$25.97/prob. & 694 min \\
  & \textbf{MerLean + Opus~4.7} & \textbf{--} & \textbf{12/12} & not reported & 789 min \\
  & \textbf{\hone{} + Opus~4.7, pass@4} & \textbf{--} & \textbf{12/12} & \$105.82/prob. & 2{,}150 min \\ %
  & \textbf{\hone{} + Opus~4.7, best@6} (cost excludes failures) & \textbf{--} & \textbf{12/12} & \$29.01/prob. & 966 min \\
\cmidrule(lr){2-2}
    & \textbf{\hone{} + Opus~4.8 / Gemini~3.1~Pro / GPT-5.5} & \textbf{--} & \textbf{12/12} & \textbf{\$7.87 / \$7.12 / \$6.84/prob.} & 792 / 608 / 278 min \\
\midrule
\multirow{4}{*}{IMO 2026}
  & AxiomProver & -- & 6/6 & not reported & 1{,}496 min \\
  & \hone{} + Gemini~3.1~Pro & -- & 6/6 & \$54.37/prob. & 2{,}005 min \\
  & Humanfia + GPT-5.6~Sol & -- & 6/6 & \$21.23/prob. & 498 min \\
  & \textbf{\hone{} + GPT-5.6~Sol ($T50$ / $\boldsymbol{T200}$)} & \textbf{--} & \textbf{6/6} & \textbf{\$8.33 / \$8.07/prob.} & 285 / 160 min \\
\midrule
\multirow{11}{*}{\shortstack[l]{FormalQualBench}}
  & Claude Code (math.inc run, Opus~4.6) & 4 h & 4/23 & \$23.16 & 1h45m \\
  & Claude Code (our rerun, Opus~4.6, \texttt{/goal}) & 4 h & 7/23 & \$17.28 & 1h34m \\
  & \textbf{OpenGauss (math.inc run, Opus~4.6)} & \textbf{4 h} & \textbf{8/23} & \textbf{\$24.93} & 1h48m \\
  & \textbf{\hone{} + Opus~4.6 (best seed of 4)} & \textbf{4 h} & \textbf{8/23} & \textbf{\$23.40} & 1h58m \\
\cmidrule(lr){2-2}
  & \hone{} + Opus~4.7 & 4 h & 8/23 & \$18.14 & 43m \\
  & MerLean + Opus~4.7 & 4 h & 9/23 & \$101.09 & 1h42m \\
  & \hone{} + Opus~4.7 & -- & 9/23 & \$43.92 & 1h41m \\
  & \textbf{MerLean + Opus~4.7} & \textbf{--} & \textbf{10/23} & \$118.35 & 1h59m \\
\cmidrule(lr){2-2}
  & \hone{} + GPT-5.6~Sol & 4 h & 11/23 & \$8.97 & 40m \\
  & \hone{} + Opus~4.8 & -- & 13/23 & \$49.46 & 5h00m \\
  & \textbf{\hone{} + GPT-5.6~Sol} & \textbf{--} & \textbf{14/23} & \textbf{\$26.38} & 1h55m \\
\bottomrule
\end{tabular}}

\end{table}

\subsection{Remaining advantages of specialized systems}
\label{app:specialized-exceptions}

Among 707 problems, two fixed-model exceptions stood out: the group-theoretic \texttt{BurnsidePrimeDegree} and the density problem Erd\H{o}s~125 (Appendix~\ref{app:erdos-individual}).
On FormalQualBench with Opus~4.7, MerLean leads \hone{} by one solve with or without the four-hour cutoff (Table~\ref{tab:complex}).
MerLean proves \texttt{BurnsidePrimeDegree} in 3h12m with a 32-node decomposition~\citep{merlean2026prover}; eleven \hone{} attempts fail despite a combined \$6{,}000 and 255 hours.
With Opus~4.8, we stopped our FormalQualBench run by hand after 17 hours because of time and compute limits (Appendix~\ref{app:budget-comparisons}); it had not proved \texttt{BurnsidePrimeDegree} by iteration 19 (\$195), so that run's 13/23 is a lower bound.
With GPT-5.6~Sol, however, \hone{} proves \texttt{BurnsidePrimeDegree} (Table~\ref{tab:complex}).
On Erd\H{o}s~125, full APN reaches 50\% at pass@1, basic reaches only 10\% at pass@10, and all sixteen \hone{} attempts end in unrecoverable Lean REPL timeouts, an interface limitation (Appendix~\ref{app:budget-comparisons}; Figure~\ref{fig:erdos-six-individual}; Appendix~\ref{app:erdos-individual}).

Specialized harnesses may still reduce time to a solution, even when \hone{} matches their solve rate at lower cost.
Numina solves twelve in less summed time than \hone{} with GPT-5.2, possibly helped by faster constituent models (Table~\ref{tab:complex}).
Multi-agent parallelization could further reduce elapsed time, a capability orthogonal to \hone{}'s simple sequential search that we discuss in Section~\ref{sec:limitations}.

\begin{figure}[H]
\centering
\includegraphics[width=0.5\linewidth]{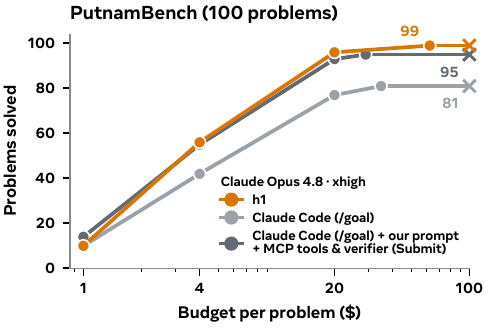}
\caption{\textbf{Claude Code improves with our Lean tools but still falls short of \hone{} on the PutnamBench subset.} With Opus~4.8, native Claude Code solves 81 problems; our prompt and Lean tools through MCP, in place of its built-in shell and file tools, raise this to 95, while \hone{} reaches 99. The FATE-X comparison appears in Figure~\ref{fig:claude-code-cap-curves}, right.}
\label{fig:claude-code-putnam}
\end{figure}

\subsection{PutnamBench: reflection and specialized provers}
\label{app:single-attempt}
\begin{figure}[H]
\centering
\includegraphics[width=0.5\linewidth]{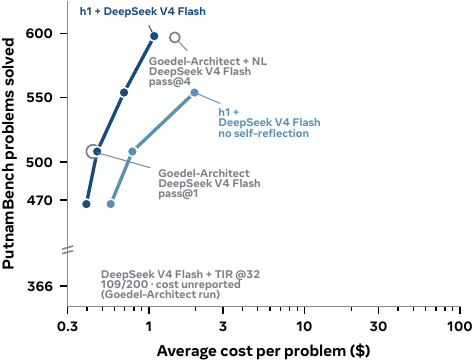}
\caption{\textbf{Reflection substantially improves \hone{}'s coverage, exceeding Goedel-Architect's pass@4 coverage at lower estimated cost.} With the same DeepSeek~V4~Flash model, \hone{} also substantially outperforms Goedel-Architect's tool-integrated reasoning (TIR) baseline, showing that its baseline understates what a simple agent can achieve.}
\label{fig:deepseek-reflection}
\end{figure}

\subsection{DeepSeek~V4~Flash on FC100-solved and FATE-X}
\label{app:deepseek-no-reflection}
\begin{figure}[H]
\centering
\begin{minipage}[t]{0.49\textwidth}
  \centering\includegraphics[width=\linewidth]{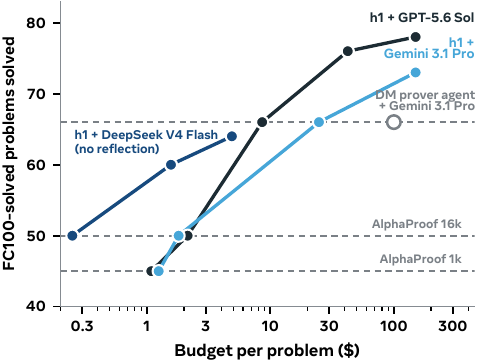}
\end{minipage}\hfill
\begin{minipage}[t]{0.49\textwidth}
  \centering\includegraphics[width=\linewidth]{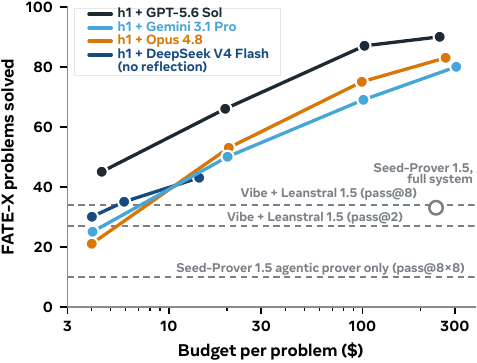}
\end{minipage}
\caption{\textbf{Figure~\ref{fig:frontiers} with the DeepSeek~V4~Flash setting labelled.}
FC100-solved (left) and FATE-X (right); the DeepSeek~V4~Flash runs use \hone{} without self-reflection.}
\label{fig:frontiers-ds-note}
\end{figure}

\begin{figure}[H]
\centering
\includegraphics[width=\linewidth]{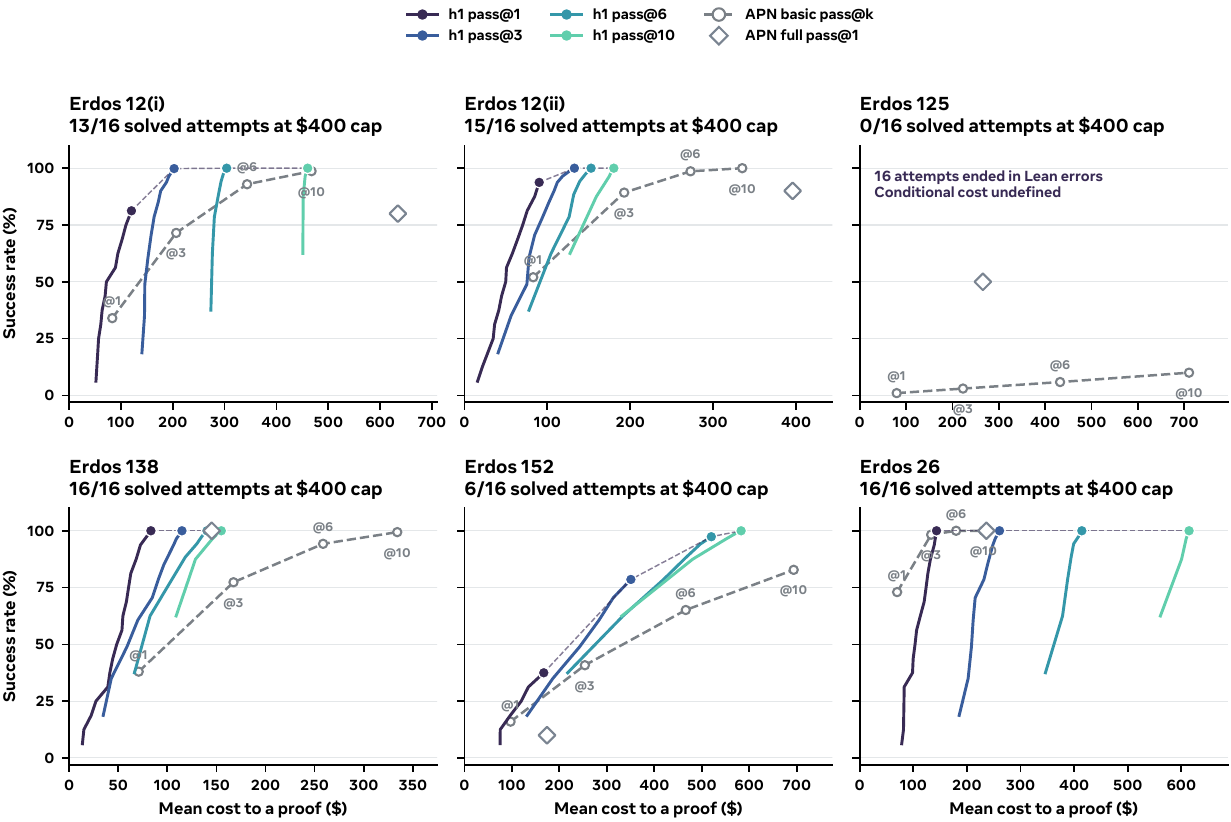}
\caption{\textbf{The five Erd\H{o}s problems in Figure~\ref{fig:teaser} (middle), plus Erd\H{o}s~125.}
Solid coloured curves vary the active-time cutoff within an attempt, separately for each
pass@$k$; filled circles mark terminal results. The dashed connection between those
circles shows repetition at the final allowance. Grey circles are APN basic's published
pass@$k$ points, and the hollow diamond is APN full's pass@1. All use Gemini~3.1~Pro.
Horizontal scales differ across panels. Following AlphaProof Nexus~\citep{alphaproofnexus2026},
we report cost conditional on success. All 16 \hone{} replicas on Erd\H{o}s~125 ended in Lean REPL timeout errors, so its
conditional cost is undefined. Errors count as failures. APN full also uses AlphaProof, whose cost is
absent from the published coordinates.}
\label{fig:erdos-six-individual}
\end{figure}

Because of a configuration error on our side, the DeepSeek~V4~Flash runs on FC100-solved and
FATE-X in Figure~\ref{fig:frontiers} ran with self-reflection disabled.
Figure~\ref{fig:frontiers-ds-note} reproduces both panels with this setting labelled.
With the same model on PutnamBench, reflection substantially improves coverage
(Figure~\ref{fig:deepseek-reflection}), so these curves may understate \hone{} with
DeepSeek~V4~Flash. We will add runs with self-reflection in a subsequent version.

\subsection{Erd\H{o}s problems: individual scaling curves}
\label{app:erdos-individual}

\subsection{Erd\H{o}s problems: prompt and model ablations}
\label{app:erdos-ablations}

\begin{figure}[H]
\centering
\includegraphics[width=0.83\linewidth]{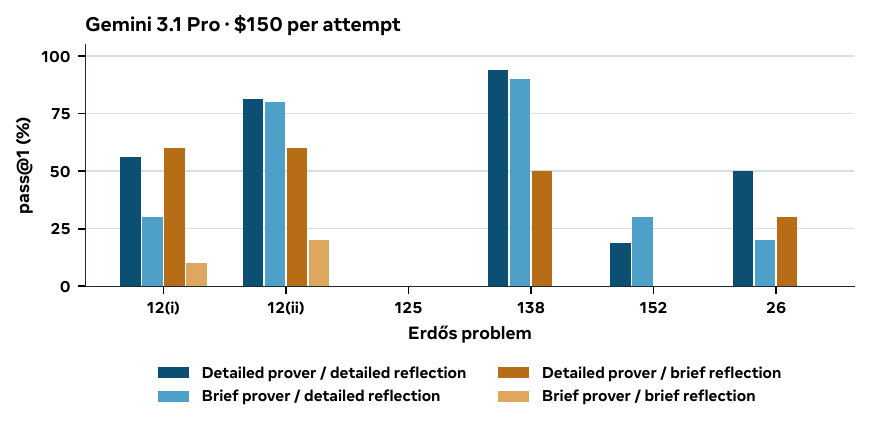}
\caption{\textbf{Per-problem success under different prompts for the two LLM roles.}
The same runs and \$150 cap as Figure~\ref{fig:erdos-prompt-main}; bars show
pass@1, with all replicas in the denominator.}
\label{fig:erdos-prompt-problems}
\end{figure}

\begin{figure}[H]
\centering
\includegraphics[width=0.83\linewidth]{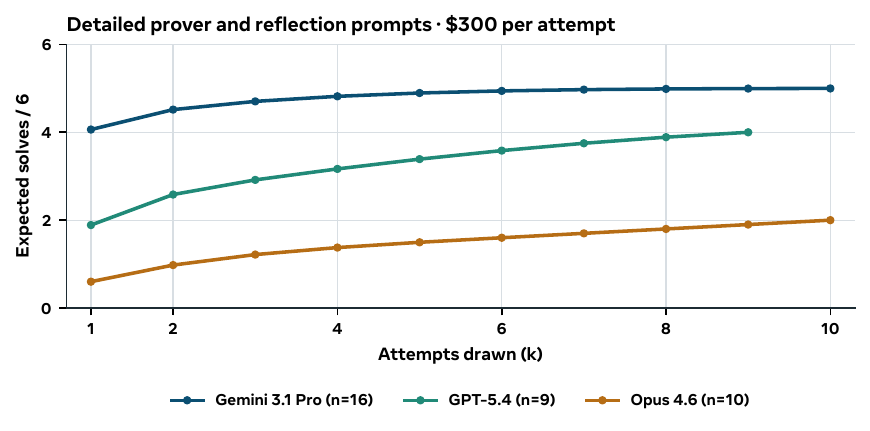}
\caption{\textbf{Model scaling on six Erd\H{o}s problems with detailed instructions.}
All models use a \$300 per-attempt cap. Curves sum per-problem pass@$k$
estimates; Gemini~3.1~Pro has 16 replicas, GPT-5.4 nine, and Opus~4.6 ten.}
\label{fig:erdos-model-curves}
\end{figure}

\begin{figure}[H]
\centering
\includegraphics[width=0.83\linewidth]{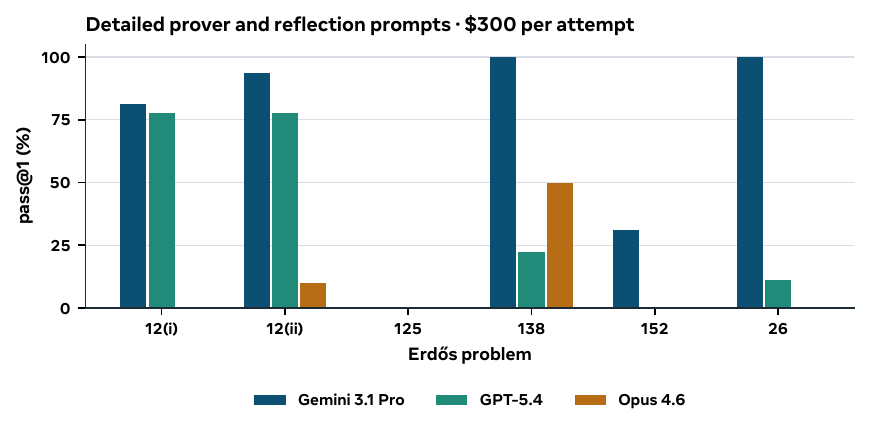}
\caption{\textbf{Per-problem success for the model comparison.}
Bars show pass@1 in the runs of Figure~\ref{fig:erdos-model-curves}, with
detailed instructions for both LLM roles and a \$300 per-attempt cap.}
\label{fig:erdos-model-problems}
\end{figure}

\subsection{Comparing success and cost with AlphaProof Nexus}
\label{app:erdos-aggregation}
We selected the six Erd\H{o}s problem instances in Figure~3 of \citet{alphaproofnexus2026} for direct comparison: 12(i), 12(ii), 26, 125, 138, and 152.

AlphaProof Nexus operates on Lean files with explicitly marked editable regions: \texttt{EVOLVE-BLOCK} regions contain proof code, while \texttt{EVOLVE-VALUE} regions allow the agent to determine values within the statement, including a conjecture's truth value~\citep[Section~2 and Appendix~A.2]{alphaproofnexus2026}.
We reconstruct our tasks from the statements and edit boundaries in APN's public GitHub release,\footnote{\url{https://github.com/google-deepmind/alphaproof-nexus-results}} removing the proofs and withholding truth values wherever these are marked editable.
This preserves the fixed parts of each statement.
We interpret these released specifications as corresponding to the tasks underlying APN's reported comparison results, although the paper does not explicitly confirm the initial files used for those experiments.

We run these experiments in an isolated environment without internet access or access to existing solutions on the host filesystem.
After observing attempts to use the Lean REPL to explore the filesystem, we also explicitly inform the model of this isolation to discourage such attempts and avoid wasting inference budget (Table~\ref{tab:reflection-failures}).
The reflection LLM also redirects the prover away from these attempts and toward mathematical work in the next iteration.
This is part of the effective repair loop that supports \hone{}'s efficiency (Section~\ref{sec:prompt-reflection}): carrying lessons from unproductive attempts into the next iteration helps direct additional inference budget toward sustained progress.

We follow \citet{alphaproofnexus2026} in reporting inference cost conditional on
success. This permits comparison with their published per-problem results. For
$k$ parallel attempts, we sum all members' costs up to the first accepted
proof, including the costs of unsuccessful members. Groups with no accepted proof
contribute to the failure rate but not to the conditional-cost mean. Thus this
metric measures the token-based cost of successful searches; it does not measure total
cost across successes and failures. APN does not report
the failed-group costs needed for this comparison.

To measure how a larger inference budget improves success at fixed $k$, we vary the
same per-attempt dollar allowance \$$B$ across problems. Within each group,
attempts stop at the first accepted proof obtained within that allowance.
For problem $i$, let $\mathcal S_{i,k}(B)$ be the successful $k$-subsets of our
$n=16$ replicas under per-attempt allowance \$$B$. If $\tau_{i,G}$ is the first
eligible proof's active-time timestamp and $C_{i,r}(\tau)$ the recorded cumulative
cost of replica $r$, we compute
\begin{equation}
 \widehat p_{i,k}(B)=\frac{|\mathcal S_{i,k}(B)|}{\binom{n}{k}},\qquad
 \widehat c_{i,k}(B)=\frac{\sum_{G\in\mathcal S_{i,k}(B)}
 \sum_{r\in G}\min\{B,C_{i,r}(\tau_{i,G})\}}{|\mathcal S_{i,k}(B)|}.
 \label{eq:erdos-conditional}
\end{equation}
Active time excludes gaps between resumed sessions. Only completed model calls
contribute cost, using the specified token prices. We average both
coordinates equally over the same fixed problem set $I$:
\begin{equation}
 \overline c_{I,k}(B)=|I|^{-1}\sum_{i\in I}\widehat c_{i,k}(B),\qquad
 \overline p_{I,k}(B)=|I|^{-1}\sum_{i\in I}\widehat p_{i,k}(B).
 \label{eq:erdos-macro}
\end{equation}
The plotted frontier retains nondominated cost--success points, using only
allowances for which every problem in $I$ has an observed success and hence a
defined conditional cost.
APN basic's pass@$k$ rates are its published grouped-run estimates; our rates
use the subset estimator over recorded replicas. APN full reports ten attempts
per problem. The published APN coordinates receive the same equal problem weights. This differs
from pooling successful attempts, which would give easier problems more weight.
The overlapping subsets do not provide independent observations for estimating
uncertainty, and APN does not report uncertainty
for this aggregate.

The cost--success frontier in Figure~\ref{fig:teaser} compares the five problems
with observed successes in all three systems: 12(i), 12(ii), 138, 152, and 26.
This subset does not include Erd\H{o}s~125: all \hone{} attempts encounter Lean timeouts,
leaving its success-conditional cost undefined. Across the full six-problem set,
including these failures, \hone{} achieves 68.75\% at pass@1, close to APN full's
71.67\%, and surpasses APN full's pass@1 success rate with 79.73\% at pass@3
(Table~\ref{tab:erdos-six-rates}).

\begin{table}[H]
\centering\small
\caption{\textbf{Success rates across all six Erd\H{o}s problems.} Each entry is
$100/6$ times the sum of the six per-problem pass@$k$ estimates, giving every
problem equal weight. For \hone{}, each estimate is
$1-\binom{16-c_i}{k}/\binom{16}{k}$, where $c_i$ is the number of accepted
replicas; all errors count as failures, including all 16 on problem~125.
APN columns average the published per-problem rates with the same weights.
These are terminal outcomes, not matched-budget estimates.}
\label{tab:erdos-six-rates}
\begin{tabular}{@{}rrrr@{}}
\toprule
$k$ & \hone{} (\%) & APN basic (\%) & APN full (\%) \\
\midrule
1 & 68.75 & 35.67 & 71.67 \\
3 & 79.73 & 63.36 & -- \\
6 & 82.90 & 76.14 & -- \\
10 & 83.33 & 81.82 & -- \\
\bottomrule
\end{tabular}

\end{table}

\paragraph{Counting unsuccessful searches.}
To compare compute allocation within \hone{}, we also report mean spend over
all groups, including those with no accepted proof. Let $\mathcal G_{i,k}$ be
all $k$-subsets of the $n=16$ replicas, and let $C_{i,r}(\infty)$ denote a
replica's recorded terminal cost. The unconditional estimate is
\begin{equation}
 \widehat u_{i,k}(B)=\frac{1}{\binom{n}{k}}\left[
 \sum_{G\in\mathcal S_{i,k}(B)}\sum_{r\in G}\min\{B,C_{i,r}(\tau_{i,G})\}
 +\sum_{G\in\mathcal G_{i,k}\setminus\mathcal S_{i,k}(B)}\sum_{r\in G}\min\{B,C_{i,r}(\infty)\}
 \right].
 \label{eq:erdos-unconditional}
\end{equation}
We use the same recorded attempts, token prices, dollar caps, and first-proof
cancellation rule as in Equation~\ref{eq:erdos-conditional}; only the averaging
includes unsuccessful groups. Each problem receives equal weight. Conditioning
on success selects different groups as $k$ changes, so the conditional metric
alone does not describe the full spending trade-off. It does not mathematically
prevent repetition from dominating a pass@1 operating point: a quick proof in one
replica can cancel expensive searches in the others. We omit APN from the
unconditional plots because its failed-group costs are unavailable.

For a hypothetical example, suppose independent attempts spend at the same constant
rate and solve at \$1 with probability $0.8$, solve at \$100 with probability $0.1$,
or fail at \$100 with probability $0.1$.
Pass@1 succeeds with probability $0.9$ at conditional mean cost
$(0.8\times1+0.1\times100)/0.9=\$12$.
Two parallel attempts, cancelled at the first proof, succeed with probability
$0.99$ at conditional mean cost $(0.96\times2+0.03\times200)/0.99=\$8$:
a quick success occurs with probability $0.96$, and a success without a quick
one with probability $0.03$.
This establishes dominance of one operating point, not of the entire pass@1 curve.

\begin{figure}[H]
\centering
\includegraphics[width=\linewidth]{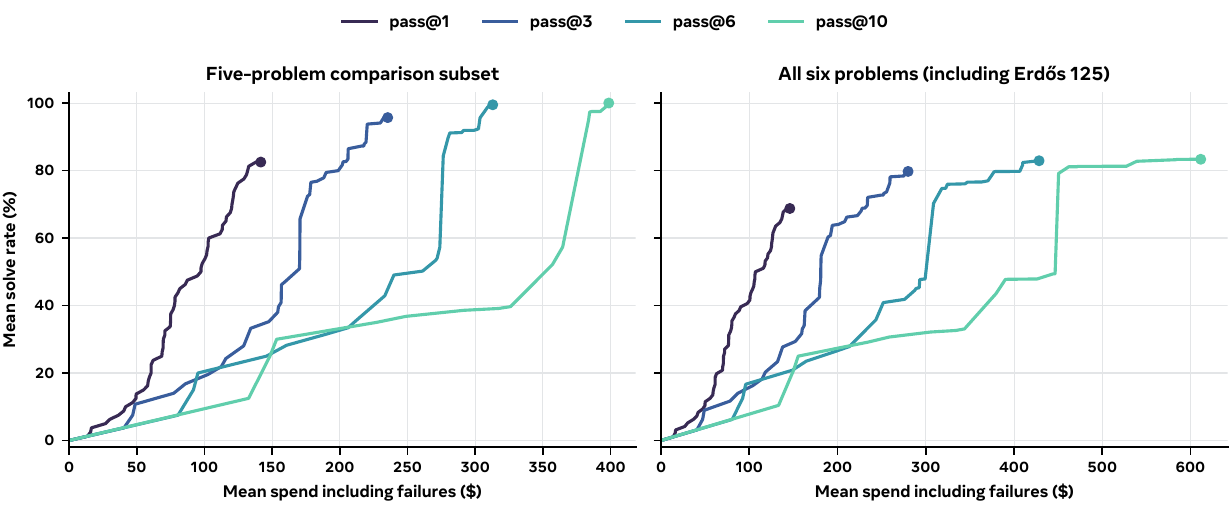}
\caption{\textbf{Longer \hone{} attempts can be cheaper than repetition even when failures are charged.}
All curves use Gemini~3.1~Pro in the same harness. Lines join nondominated
observed points as the per-attempt dollar cap varies; circles mark terminal
results. Cost averages successful and unsuccessful groups, with cancellation
at the first accepted proof in successful groups. Left: the five-problem
subset used in Figure~\ref{fig:teaser}. Right: all six problems, including
Erd\H{o}s~125 with zero accepted attempts and its recorded costs.
Its unconditional mean cost remains defined even with no successful attempts,
whereas its conditional cost is undefined.
Solve rates and costs give each problem equal weight.}
\label{fig:erdos-unconditional-aggregate}
\end{figure}

\begin{figure}[H]
\centering
\includegraphics[width=\linewidth]{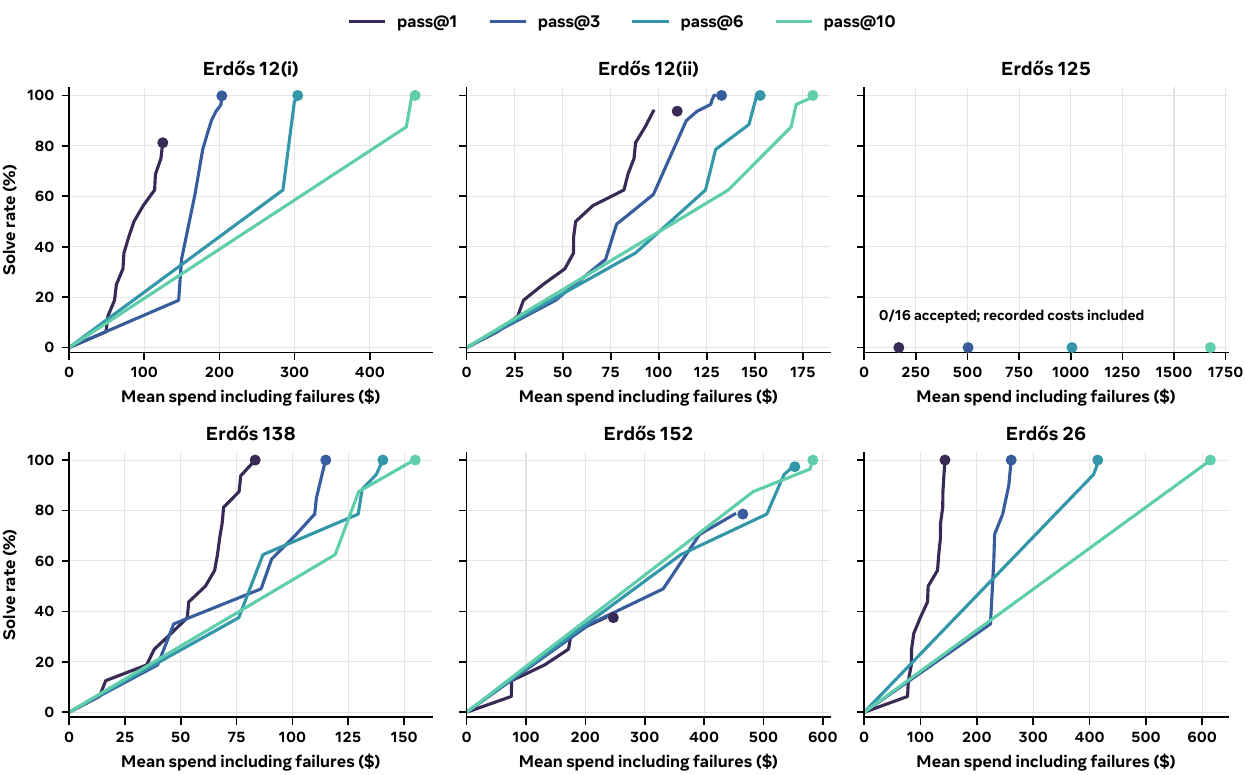}
\caption{\textbf{The trade-off between longer attempts and repetition varies by problem.}
The same unconditional-cost calculation as Figure~\ref{fig:erdos-unconditional-aggregate},
shown separately for each of the six Erd\H{o}s problems. Lines join nondominated
observed points; circles mark terminal results. All groups contribute to cost,
including those that terminate in errors. Horizontal scales differ across panels.
On Erd\H{o}s~152, pass@10 reaches 87.5\% at \$483 and leads the tested smaller
values of $k$ until pass@6 reaches 94.2\% at \$535.
 }
\label{fig:erdos-unconditional-individual}
\end{figure}

\begin{figure}[H]
\centering
\includegraphics[width=\textwidth]{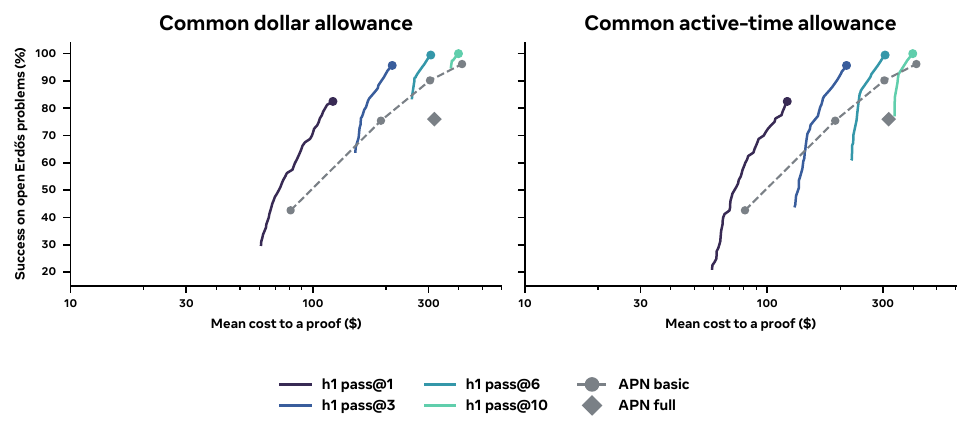}
\caption{\textbf{Dollar and time alignment give the same terminal comparisons.}
Both panels average the same five problems equally and count each group's cost
up to its first eligible proof. Only the common allowance across problems
changes. Solid curves vary that allowance at fixed $k$; circles mark terminal
results. APN's published points are identical in both panels.}
\label{fig:erdos-alignment}
\end{figure}

Figure~\ref{fig:erdos-alignment} compares using a common dollar allowance with
using a common active-time allowance across problems. Both give the same final
costs and success rates, but different intermediate curves.

\subsection{Model choice at different budgets}
\label{app:model-cost-frontier}
\begin{figure}[H]
\centering
\includegraphics[width=0.58\linewidth]{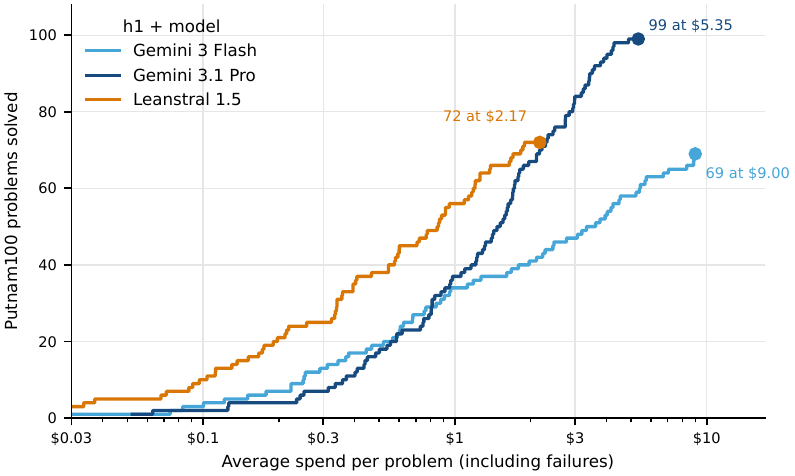}
\caption{\textbf{The most economical model depends on the target solve rate.}
On the PutnamBench subset, Leanstral~1.5 supplies the frontier through 72 solves and
Gemini~3.1~Pro extends it to 99, while Gemini~3~Flash lies below their combined
frontier. Each curve varies the per-problem cap; cost averages all 100 problems,
including failures. Dots mark full-run endpoints. Leanstral costs are
API-equivalent estimates using a measured 65\% average cache rate
at maximum context length.}
\label{fig:model-cost-frontier}
\end{figure}

\subsection{Additional comparisons with AxProverBase and analysis of failed attempts}
\label{app:harness-comparisons}
\label{app:flash-harness-traces}
\begin{figure}[H]
\centering
\includegraphics[width=\linewidth,height=0.42\textheight,keepaspectratio]{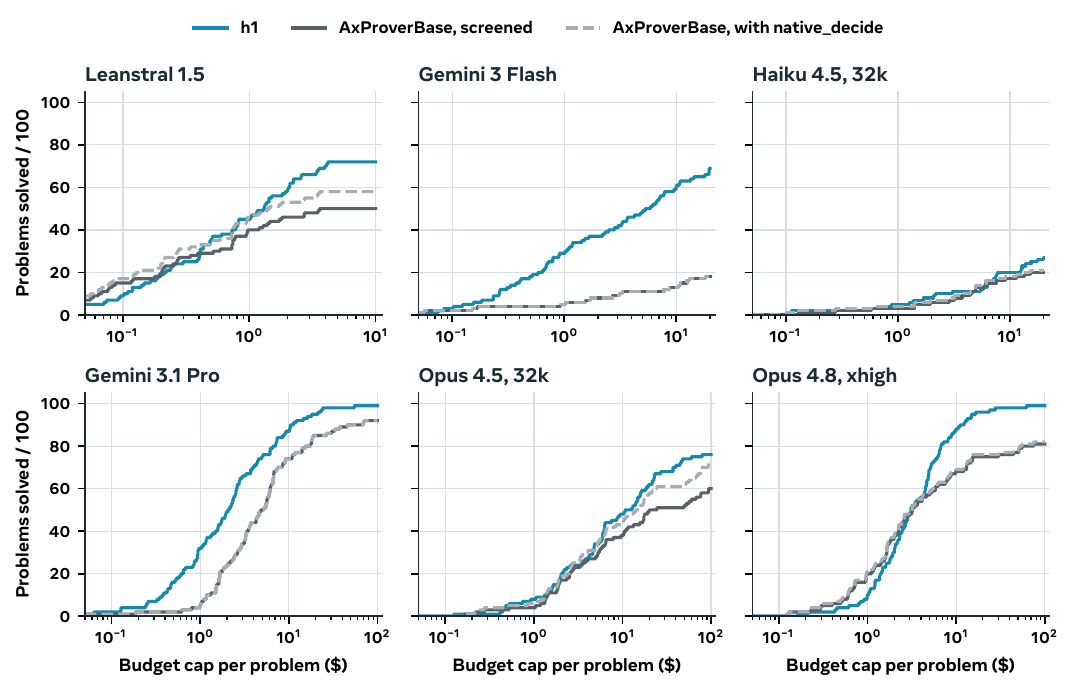}
\caption{\textbf{\hone{} overtakes AxProverBase as the budget grows across the six settings in Figure~\ref{fig:putnam100-pair}.} AxProverBase is shown both before and after screening out proofs using
\texttt{native\_decide}; overlapping lines indicate no screened solves.
All \hone{} solves pass external validation.}
\label{fig:harness-curves}
\end{figure}

\begin{figure}[H]
\centering
\includegraphics[width=0.85\linewidth]{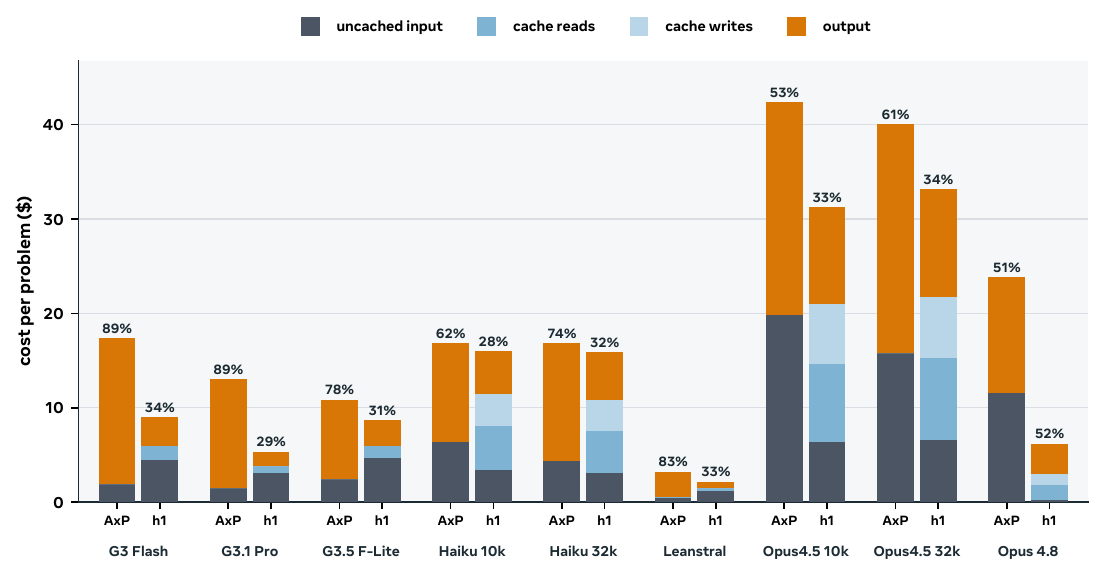}
\caption{\textbf{Cost per problem by token type, PutnamBench subset, nine model settings under AxProverBase (AxP) and \hone{}.} Bars average costs across solved and unsolved problems. Percentages show output tokens' share of total cost.}
\label{fig:cost-breakdown}
\end{figure}

With Gemini~3~Flash on the PutnamBench subset, \hone{} solves all 18 problems solved by
AxProverBase and 51 additional problems. On those 51, mean proposer-call counts are 287 for AxProverBase and 275 for \hone{}.
Mean output per problem, including reflection, is 5.97M versus 0.628M tokens;
mean input is 4.24M versus 23.13M, respectively. The fraction of input served
from cache is 0.04\% for AxProverBase and 78.8\% for \hone{}. AxProverBase regenerates a complete
candidate after feedback; \hone{} retains the file and makes local edits. These
observations connect the token-cost difference to how the systems preserve work,
without isolating it from their other design differences.

\begin{figure}[H]
\centering
\includegraphics[width=0.55\linewidth]{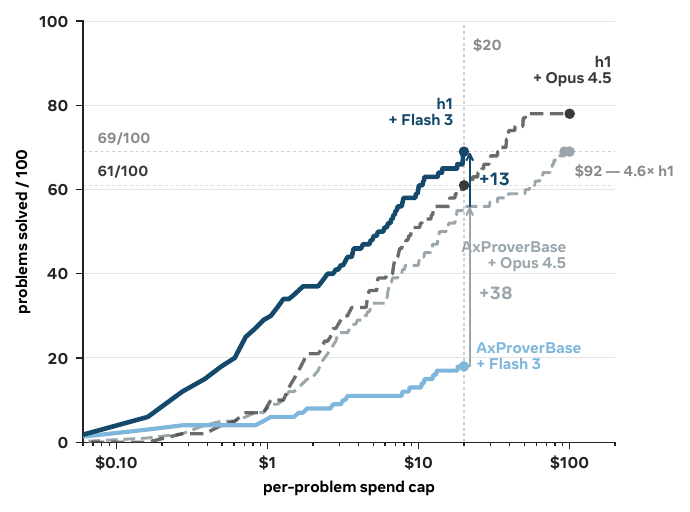}
\caption{\textbf{PutnamBench subset solve rates for \hone{} and AxProverBase as the budget increases.}
\hone{} with Opus~4.5 solves 61 problems at the \$20 per-problem cap.}
\label{fig:comparison-opus}
\end{figure}

\begin{figure}[H]
\centering
\includegraphics[width=0.88\linewidth]{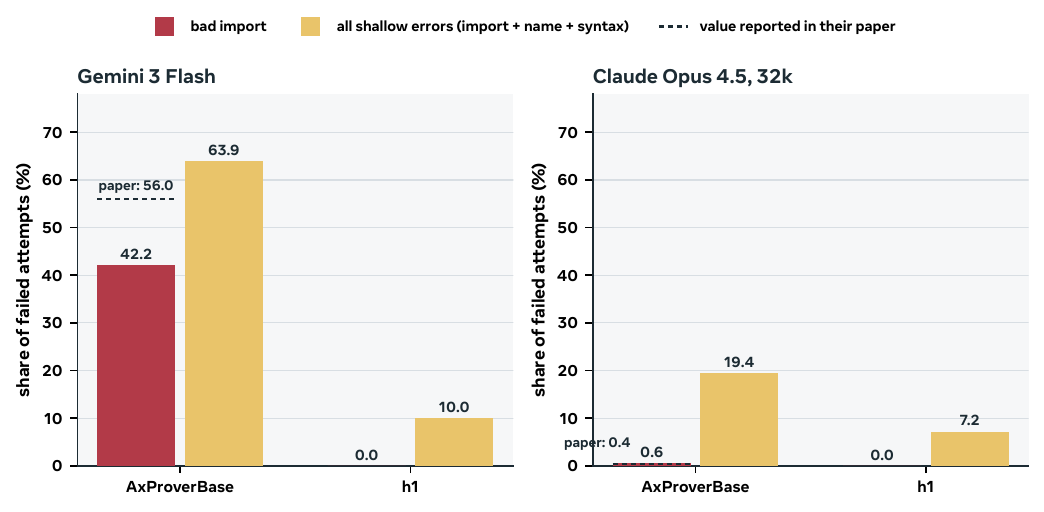}
\caption{\textbf{Bad imports consume proof-checking attempts.} The fraction of
failed builds whose first Lean error is a bad import is 42.2\% for AxProverBase
with Gemini~3~Flash and 0.6\% with Opus~4.5; dashed lines show the paper's
reported 56\% and 0.4\%. \hone{} seeds the working file with \texttt{import Mathlib}
and records no such failures.}
\label{fig:comparison-extra}
\end{figure}

For example, on \texttt{putnam\_1972\_b4}, AxProverBase uses tokens equivalent to \$20.01 over
287 builds, including 175 bad-import and 84 syntax failures. \hone{} solves it at an API-equivalent cost of
\$0.15 with 20 local edits, three REPL tests, and one submission. Across the 51 \hone{}-only solves,
AxProverBase performs 13{,}446 reflections and \hone{} performs 316, while the
systems differ in whether they regenerate or edit the working file.

\subsection{Additional comparisons with Claude Code and analysis of failed attempts}
\label{app:claude-code-mechanisms}
With the same Claude model, native Claude Code (with its built-in shell and file tools), Claude Code with our prompt and Lean tools through MCP (which replace the built-in tools),
and \hone{} solve 81/95/99 PutnamBench subset problems and 63/73/83 FATE-X problems,
respectively (Figures~\ref{fig:claude-code-putnam} and~\ref{fig:claude-code-cap-curves}, right). The tools expose Lean
feedback and SafeVerify rejection within the session; \hone{} additionally enforces
validation and another repair round after rejection.

Relative to native Claude Code, exposing final-proof validation addresses a
frequent failure mode. Of 30 problems solved by \hone{}
where native Claude Code stops normally without an accepted proof, 28 end with
an artifact rejected by SafeVerify. Adding our Lean tools to Claude Code solves
20 of these problems. \hone{} uses multiple rounds on 21 of the 30.

Relative to Claude Code with MCP tools, \hone{} additionally controls how execution
continues after failures. On FATE-X problem~19, Claude Code with our tools
resubmits an unchanged file 170 times, whereas \hone{} edits after
each of 17 negative outcomes and succeeds.

The MCP integration also leaves API-call retries under Claude Code's control.
Some Claude Code runs fail through API-call errors and repeated
timeouts. We took care to give these baselines their strongest showing: we restarted
affected runs from scratch, and reran configurations whose wall-time limit was binding
with a larger allowance. Exposing
Lean tools through MCP does not give the experimenter the same control over
Claude Code's internal retries and failure handling as implementing the harness
directly. In \hone{}, these policies are explicit: transient connection failures can
retry the same call, while verifier failures return feedback for further edits.

\subsection{Model and benchmark release dates}
\label{app:chronology}
\begin{figure}[H]
\centering
\includegraphics[width=0.85\linewidth]{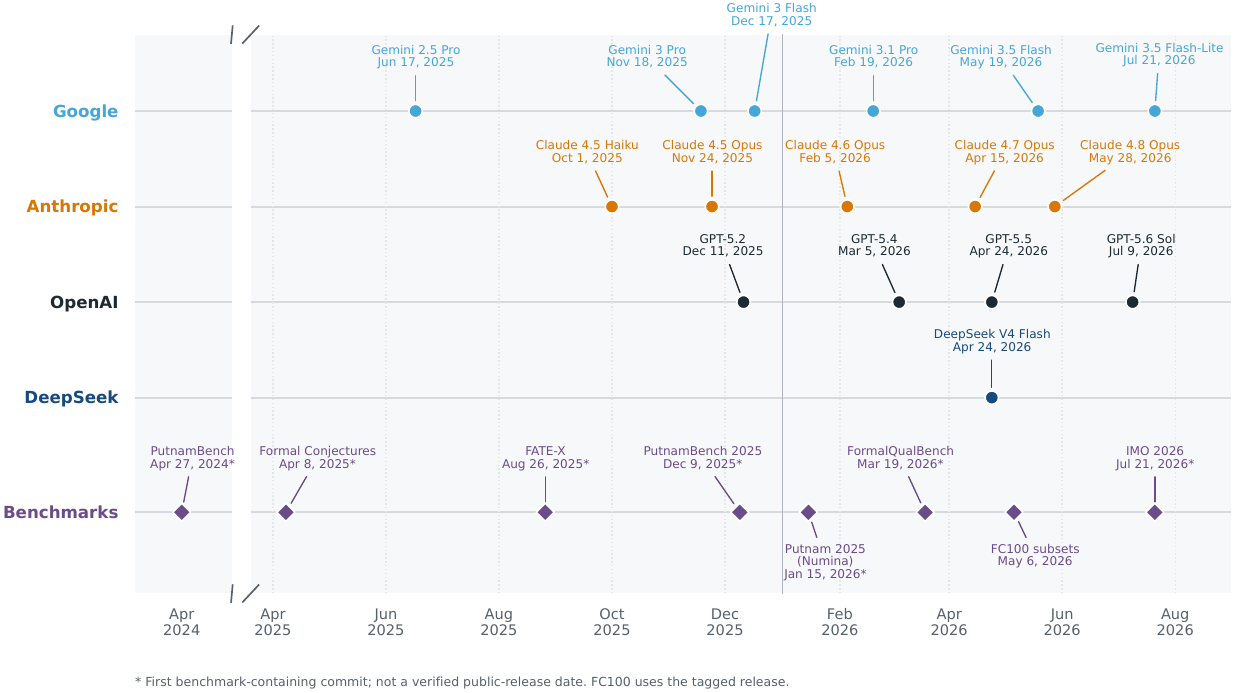}
\caption{\textbf{Model releases and benchmark dates.} Benchmark dates marked $^*$ are first benchmark-containing commits, not verified public-release dates; FC100 uses the tagged benchmark release. Dates distinguish possible exposure to statements, formalizations, and published solutions; they do not establish the contents of proprietary training data.}
\label{fig:timeline}
\end{figure}

\clearpage

\section{Tool, Prompt, and Reflection Ablations}
\label{app:ablation-extra}
\label{app:ablation-settings}

\begin{figure}[H]
\centering
\includegraphics[width=0.75\linewidth]{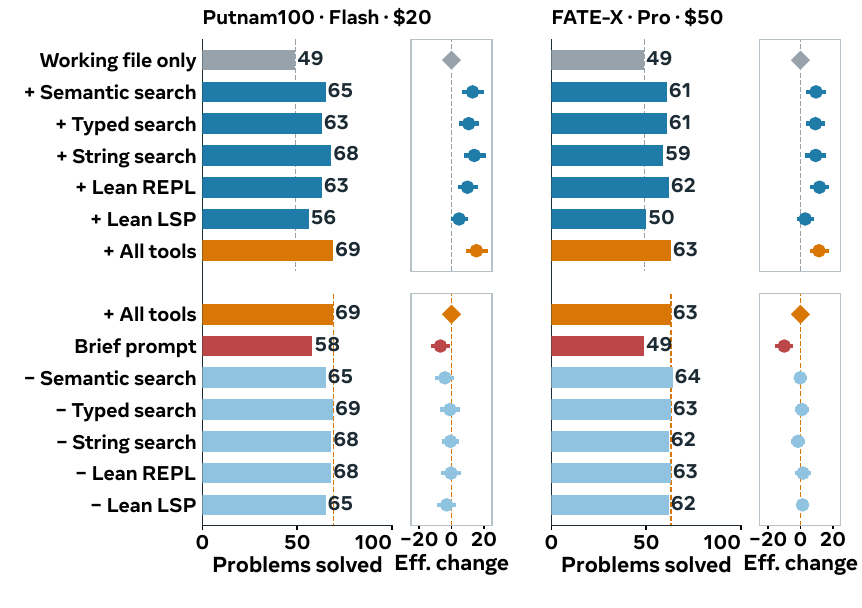}
\caption{\textbf{Tools provide overlapping benefits.} Individual search tools or the REPL recover most of the full tool set's gains; removing any single tool has little effect on coverage. Results cover the PutnamBench subset with Gemini~3~Flash and FATE-X with Gemini~3.1~Pro.}
\label{fig:tools-ablation}
\end{figure}

\begin{figure}[H]
\centering
\includegraphics[width=0.5\linewidth]{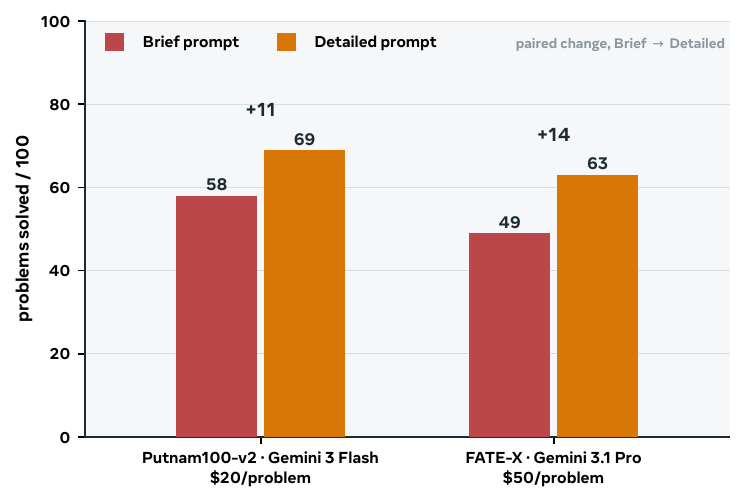}
\caption{\textbf{The detailed prompt improves solve rates on both ablation settings.}
Gemini~3~Flash on the PutnamBench subset (\$20 per problem) and Gemini~3.1~Pro on
FATE-X (\$50); all 100 problems per arm remain in the denominator, including crashes.
The prompt variants are specified in Appendix~\ref{app:prompts}.}
\label{fig:prompt}
\end{figure}

The reflection-input ablation varies the context supplied to the reflection LLM,
independently of its system prompt. With the transcript, it receives the latest
attempt's reasoning, final code, verifier feedback, and interaction transcript.
The distilled final attempt consists of the multi-turn LLM's account of its
latest attempt, the final Lean code, the verifier feedback, and the previous
reflection summary; the sequence of intermediate model turns, tool calls, and
tool results is omitted.

\begin{figure}[H]
\centering
\includegraphics[width=0.5\linewidth]{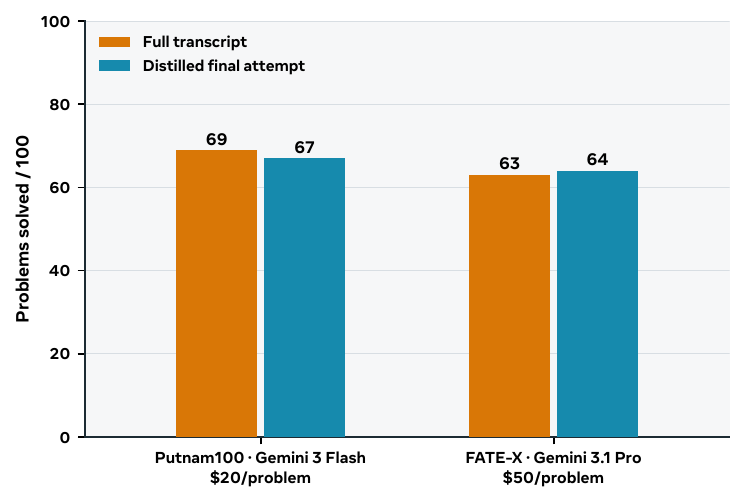}
\caption{\textbf{The two reflection LLM inputs yield similar solve counts in these runs.}
Full transcript versus distilled final attempt: 69 versus 67 solves on the PutnamBench subset
with Gemini~3~Flash, and 63 versus 64 on FATE-X with Gemini~3.1~Pro.
The per-problem caps are \$20 and \$50, respectively, including prover and reflection LLM calls; each arm contains 100 problems and one run.}
\label{fig:reflector-input}
\end{figure}

\begin{table}[H]
\centering\scriptsize
\setlength{\tabcolsep}{3pt}
\caption{\textbf{Lower input costs for reflection are offset by higher prover-call costs.}
Mean dollars per problem with full transcript versus distilled final attempt.
Relative changes compare mean costs; brackets give 95\% percentile intervals from
50,000 paired bootstrap samples of problems.}
\label{tab:reflector-costs}
\begin{tabular}{@{}lrrrrrr@{}}
\toprule
& \multicolumn{3}{c}{Putnam100} & \multicolumn{3}{c}{FATE-X} \\
\cmidrule(lr){2-4}\cmidrule(lr){5-7}
Cost component & Full & Distilled & Change [95\% CI] & Full & Distilled & Change [95\% CI] \\
\midrule
Reflector input & 0.91 & 0.04 & $-95.1\%$ [$-95.6, -94.6$] & 1.74 & 0.18 & $-89.6\%$ [$-90.9, -88.2$] \\
Reflector output & 0.10 & 0.38 & $+260.4\%$ [$+217.7, +311.1$] & 0.92 & 0.69 & $-24.2\%$ [$-35.5, -11.5$] \\
Prover calls & 7.99 & 8.75 & $+9.6\%$ [$-0.1, +20.0$] & 23.37 & 25.09 & $+7.3\%$ [$+1.1, +14.3$] \\
Total & 9.00 & 9.17 & $+1.9\%$ [$-6.9, +11.5$] & 26.02 & 25.96 & $-0.2\%$ [$-6.0, +6.1$] \\
\bottomrule
\end{tabular}

\end{table}

Removing the transcript reduces input costs for reflection by 95\% on the PutnamBench subset and
90\% on FATE-X, but prover-call costs rise by 9.6\% and 7.3\%, respectively
(Table~\ref{tab:reflector-costs}). The increase occurs on 64/100 and 74/100 problems.
Mean prover-call counts rise by 12.1\% [1.4, 23.5] and 18.3\% [11.1, 26.2].
Total-cost changes remain small; intervals span zero and reflect variation across paired problems.

\begin{figure}[H]
\centering
\includegraphics[width=\linewidth,height=0.28\textheight,keepaspectratio]{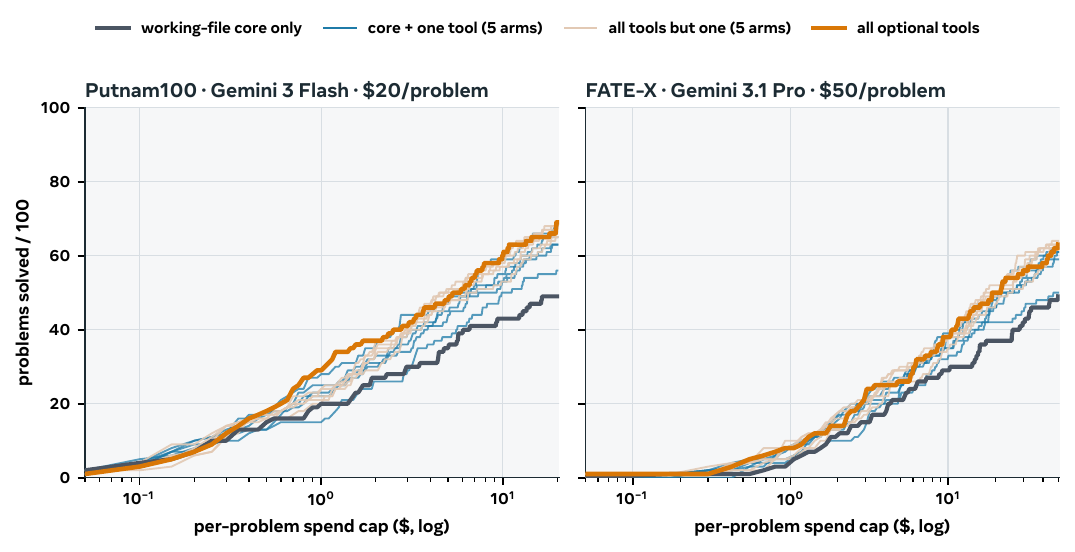}
\caption{\textbf{Solve rates under different tool configurations and inference budgets.}
Gemini~3~Flash on the PutnamBench subset (\$20 per problem; left) and Gemini~3.1~Pro on
FATE-X (\$50; right). The working-file baseline permits editing and checking
the proof file. Blue curves add one of five capabilities: semantic, type, or
text search, the Lean REPL, or the language server. Pale curves remove one
capability from the complete tool set. The working-file baseline generally
solves fewer problems; the other configurations substantially overlap.
On the PutnamBench subset, all tools retain all 49 working-file-baseline solves and add 20;
on FATE-X, they retain 48 of 49 and add 15.}
\label{fig:tools-curves}
\end{figure}

\label{app:substitution}

\begin{figure}[H]
\centering
\includegraphics[width=0.92\linewidth]{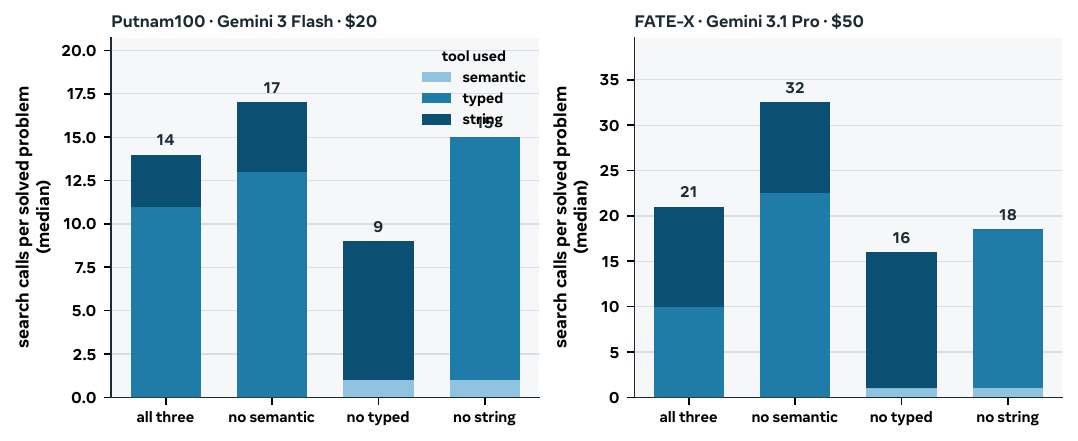}\par\medskip
\includegraphics[width=0.58\linewidth]{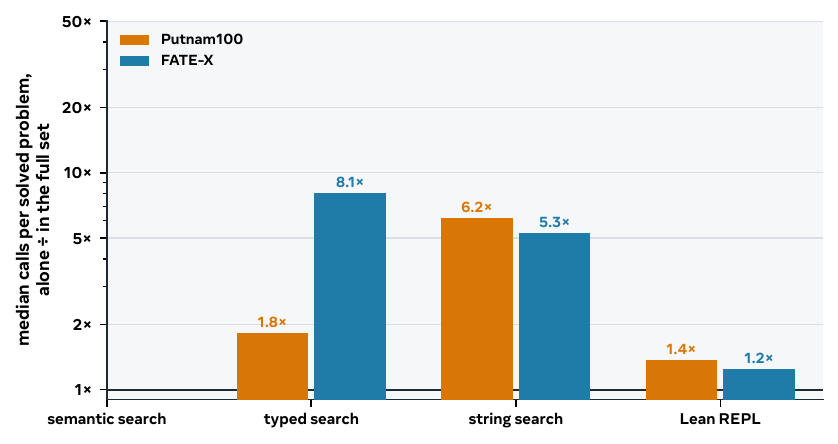}
\caption{\textbf{Tool use when search and execution capabilities change.}
Top: median calls to each search tool among problems solved by each configuration;
stacked heights sum these tool-specific medians. Bottom: each tool's median
call count when it is the only additional capability, divided by its median
with all tools available. These comparisons use each configuration's own solved
problems.}
\label{fig:substitution}
\end{figure}

\begin{figure}[H]
\centering
\includegraphics[width=\linewidth,height=0.28\textheight,keepaspectratio]{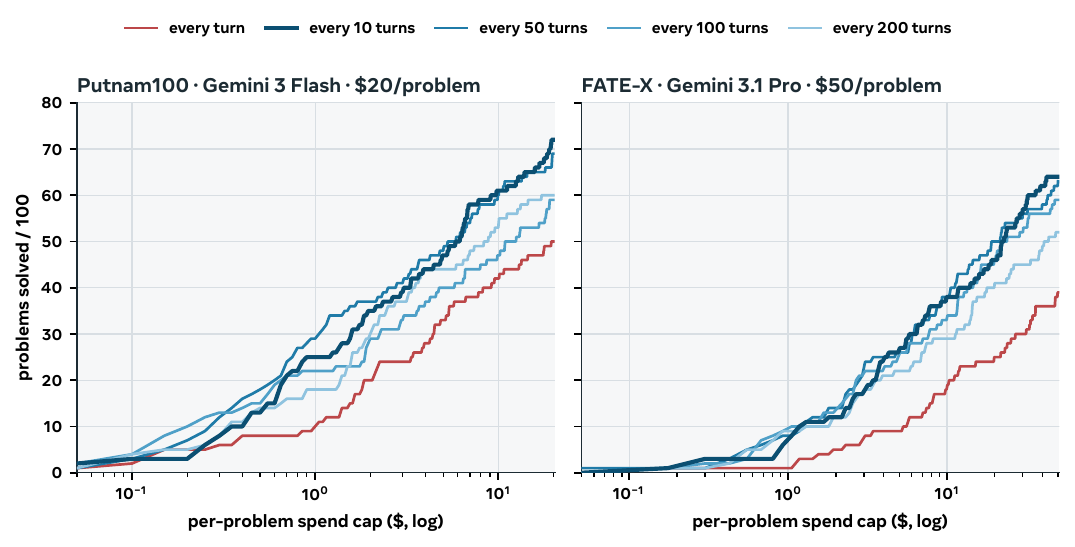}
\caption{\textbf{Solve rates at different budgets and reflection intervals.}
Reflection follows every 1, 10, 50, 100, or 200 proof-search turns ($T1$--$T200$).
Gemini~3~Flash is evaluated on the PutnamBench subset at up to \$20 per problem (left), and
Gemini~3.1~Pro on FATE-X at up to \$50 (right). Reflection every turn produces
fewer solves over most observed budgets; intervals of 10 and 50 turns give
similar curves.}
\label{fig:cadence-curves}
\end{figure}

\begin{figure}[H]
\centering
\includegraphics[width=0.65\linewidth]{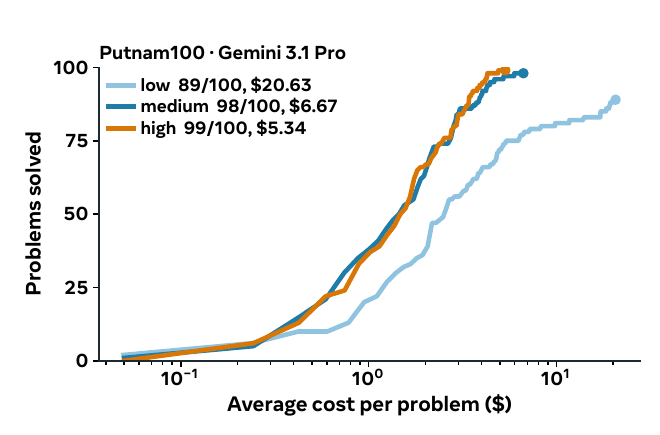}
\caption{\textbf{More reasoning per turn can make test-time scaling more efficient.}
Gemini~3.1~Pro on the PutnamBench subset with low, medium, and high reasoning effort.
Curves plot solves against mean cost, including failed attempts;
legends report final solve counts and mean costs.}
\label{fig:reasoning-effort}
\end{figure}

\begin{figure}[H]
\centering
\includegraphics[width=\linewidth]{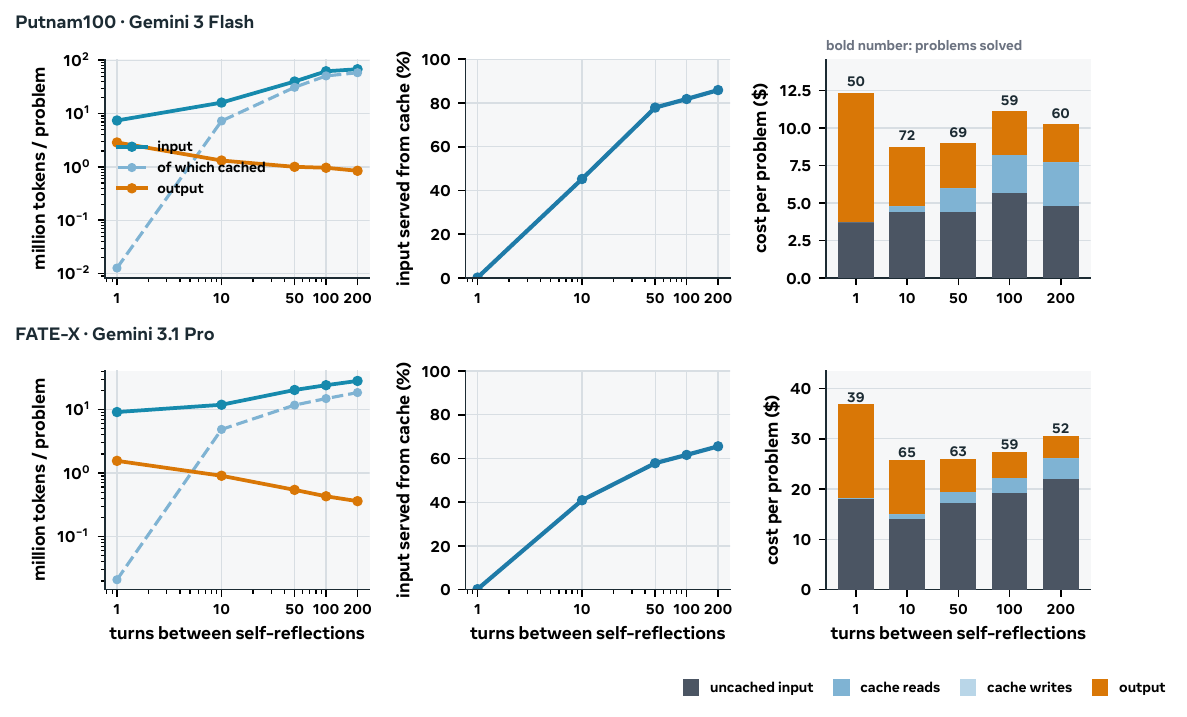}
\caption{\textbf{Token use and cost at different reflection intervals.}
Gemini~3~Flash on the PutnamBench subset (top) and Gemini~3.1~Pro on FATE-X (bottom).
Left: input, cached input, and output tokens averaged over recorded problems.
Middle: the fraction of all input tokens served from cache. Right: mean inference
cost per problem, including failed attempts, split by token type; numbers above
bars give solve counts. Calls to both the multi-turn LLM and reflection LLM are included.
The horizontal axis gives the maximum number of proof-search turns between reflections.}
\label{fig:effort-econ}
\end{figure}

\clearpage

\section{Solve-Time and Proof-Size Statistics}
\label{app:search-horizon}

The figures below cover the four state-of-the-art runs. Across all our runs, the maxima reported in Section~\ref{sec:benchmarks} come from other runs: 1{,}123 iterations before acceptance on PutnamBench problem \texttt{putnam\_1992\_b3}; 95 active hours for DeepSeek~V4~Flash on FATE-X problem~38, accepted at iteration 546; and an accepted file of 7{,}124 lines for the Borsuk--Ulam theorem.

\begin{figure}[H]
\centering
\includegraphics[width=\linewidth]{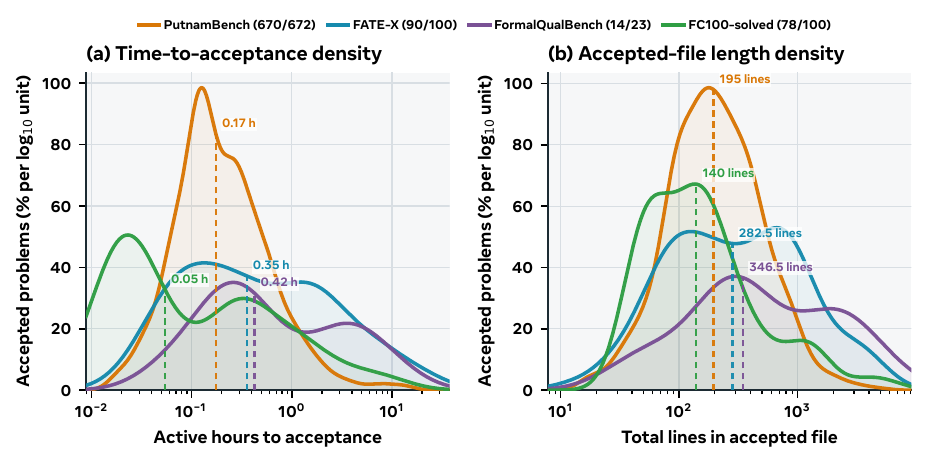}
\caption{\textbf{Time to acceptance and accepted-file length.}
The selected runs contain 670/672 accepted PutnamBench problems with GPT-5.5,
and 90/100 on FATE-X, 14/23 on FormalQualBench, and 78/100 on FC100-solved
with GPT-5.6~Sol: 852 accepted proofs in 895 attempts. Left: active hours to
first acceptance. Right: total lines in the accepted Lean file. Densities
are estimated in $\log_{10}$ space and normalized so their area in that space
equals the benchmark's acceptance percentage. Dashed lines mark medians among
accepted proofs.}
\label{fig:search-horizon-distributions}
\end{figure}

\begin{figure}[H]
\centering
\includegraphics[width=\linewidth]{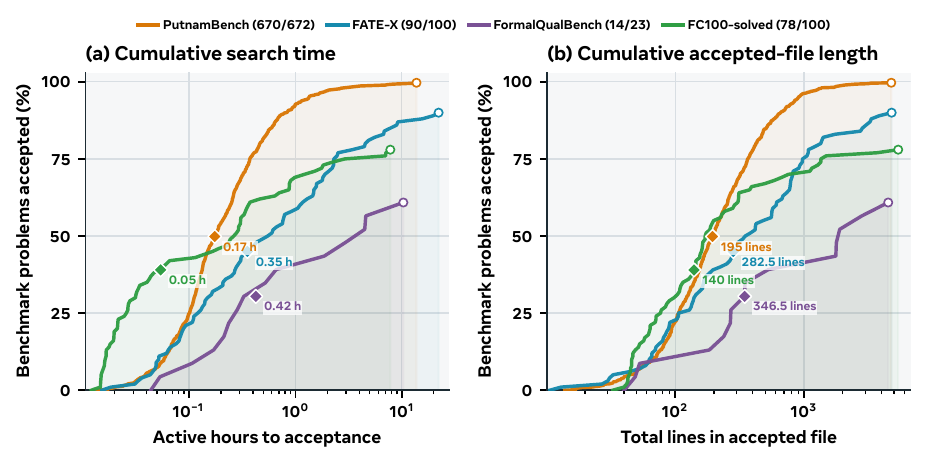}
\caption{\textbf{Cumulative distributions of time to acceptance and accepted-file length.}
The runs are those in Figure~\ref{fig:search-horizon-distributions}. Counts are
divided by all attempted problems, so each curve ends at the run's acceptance
percentage. Diamonds mark medians among accepted proofs.}
\label{fig:search-horizon-cumulative}
\end{figure}

\begin{figure}[H]
\centering
\includegraphics[width=0.82\linewidth]{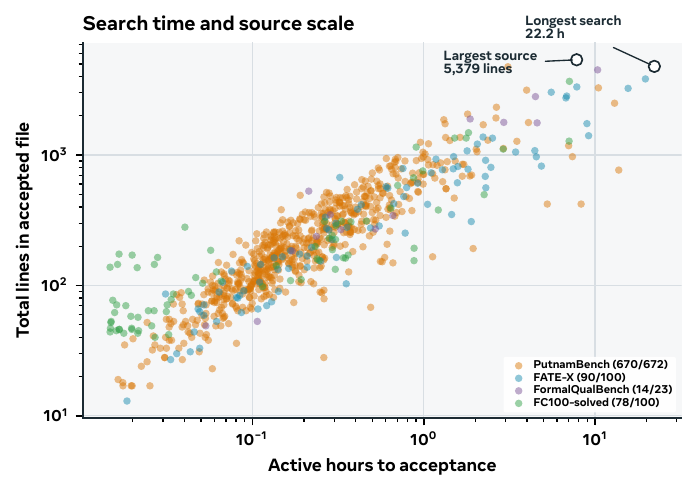}
\caption{\textbf{Search time and total file length across 852 accepted proofs.}
GPT-5.5 on PutnamBench; GPT-5.6~Sol on FATE-X, FormalQualBench, and FC100-solved.}
\label{fig:search-time-source-scale}
\end{figure}

\clearpage
\section{Reproducibility}
\label{app:reproducibility}

We have included as much detail as possible in the paper and appendices to support
reproduction of both the harness and the procedures used to plot the figures.
Section~\ref{sec:harness} and Appendix~\ref{app:harness} describe the harness, including
its tools, prompts, and reflection protocol. Appendix~\ref{app:evaluation} documents
formal validation, inference budgets, pricing, and cost-capped curve construction;
Appendix~\ref{app:erdos-aggregation} details aggregation and uncertainty estimates;
and Appendix~\ref{app:ablation-settings} specifies the ablation settings.

\clearpage
\raggedbottom
\end{document}